\documentclass[final,12pt, 3p,times]{elsarticle}

\usepackage{amsmath} 
\usepackage{color}
\usepackage{balance} 

\usepackage{arydshln}
\usepackage{booktabs}
\usepackage{adjustbox}
\usepackage[ruled, linesnumbered]{algorithm2e}
\usepackage{bbm}
\usepackage{bm}
\usepackage{framed} 
\usepackage{graphicx}
\usepackage{graphics}
\usepackage{multirow}
\usepackage{float}
\usepackage{microtype}      
\usepackage{rotating}
\usepackage{subcaption}

\usepackage{breakcites}
\usepackage[toc,page]{appendix}

\usepackage{hyperref}

\usepackage{amsthm} 

\usepackage[capitalize]{cleveref}
\crefname{section}{Sec.}{Secs.}
\Crefname{section}{Section}{Sections}
\Crefname{table}{Table}{Tables}
\crefname{table}{Tab.}{Tabs.}

\newcommand\scalemath[2]{\scalebox{#1}{\mbox{\ensuremath{\displaystyle #2}}}}  

\def\rev#1{\textcolor{black}{#1}}

\newtheorem{theorem}{Theorem}

\usepackage{amsthm}

\newtheorem{remark}{Remark}

\AtBeginDocument{%
  \providecommand\BibTeX{{%
    \normalfont B\kern-0.5em{\scshape i\kern-0.25em b}\kern-0.8em\TeX}}}

\usepackage{lineno}

\journal{XXX}

\begin{document}
\date{}
\begin{frontmatter}




\title{Efficient Subsampling of Realistic Images From GANs \\ Conditional on a Class or a Continuous Variable}

\author[contact1]{Xin Ding}
\ead{dingx92ubc@126.com}
\author[contact2]{Yongwei Wang\corref{cor1}}
\ead{yongweiw@ece.ubc.ca}
\author[contact2]{Z. Jane Wang}
\ead{zjanew@ece.ubc.ca}
\author[contact2]{William J. Welch}
\ead{will@stat.ubc.ca}

\cortext[cor1]{Corresponding author.}

\address[contact1]{Department of Statistics, University of British Columbia, Vancouver, BC, V6T 1Z4, Canada\\}
\address[contact2]{Department
	of Electrical and Computer Engineering, University of British Columbia, Vancouver,
	BC, V6T 1Z4, Canada\\}

\begin{abstract}
Recently, subsampling or refining images generated from unconditional generative adversarial networks (GANs) has been actively studied to improve the overall image quality. Unfortunately, these methods are often observed to be less effective or inefficient in handling conditional GANs (cGANs) --- conditioning on a class (a.k.a class-conditional GANs) or a continuous variable (a.k.a continuous cGANs or CcGANs). In this work, we introduce an effective and efficient subsampling scheme, named \textbf{c}onditional \textbf{D}ensity \textbf{R}atio-guided \textbf{R}ejection \textbf{S}ampling (cDR-RS), to sample high-quality images from cGANs. Specifically, we first develop a novel conditional density ratio estimation method, termed cDRE-F-cSP, by proposing an improved feature (F) extraction mechanism and a conditional Softplus (cSP) loss. We then derive the error bound of a density ratio model trained with the cSP loss. Finally, we accept or reject a fake image in terms of its estimated conditional density ratio. A filtering scheme is also developed to increase fake images' label consistency without losing diversity when sampling from CcGANs. We extensively test the effectiveness and efficiency of cDR-RS in sampling from both class-conditional GANs and CcGANs on five benchmark datasets. This comparison includes state-of-the-art subsampling or refining methods (e.g., DRS, Collab, DDLS, DRE-F-SP+RS). \textit{When sampling from class-conditional GANs}, cDR-RS outperforms all state-of-the-art methods (except DRE-F-SP+RS) by a large margin in terms of effectiveness (i.e., Intra-FID, FID, and IS scores). Although the effectiveness of cDR-RS is often comparable to that of DRE-F-SP+RS, cDR-RS is substantially more efficient. For example, cDR-RS only requires \textbf{1.19\%} of the storage usage and \textbf{77\%} of the implementation time spent by DRE-F-SP+RS on ImageNet-100. \textit{When sampling from CcGANs}, the superiority of cDR-RS is even more noticeable in terms of both effectiveness and efficiency. Notably, with the consumption of reasonable computational resources, cDR-RS can substantially reduce the Label Score metric without decreasing the diversity of CcGAN-generated images, while other methods often need to trade much diversity for slightly improved Label Score.
\end{abstract}

\begin{keyword}
conditional generative adversarial networks \sep subsampling \sep conditional density ratio estimation
\end{keyword}

\end{frontmatter}

\section{Introduction}\label{sec:intro}

\begin{figure}[ht]
	\centering
	\includegraphics[width=1\textwidth]{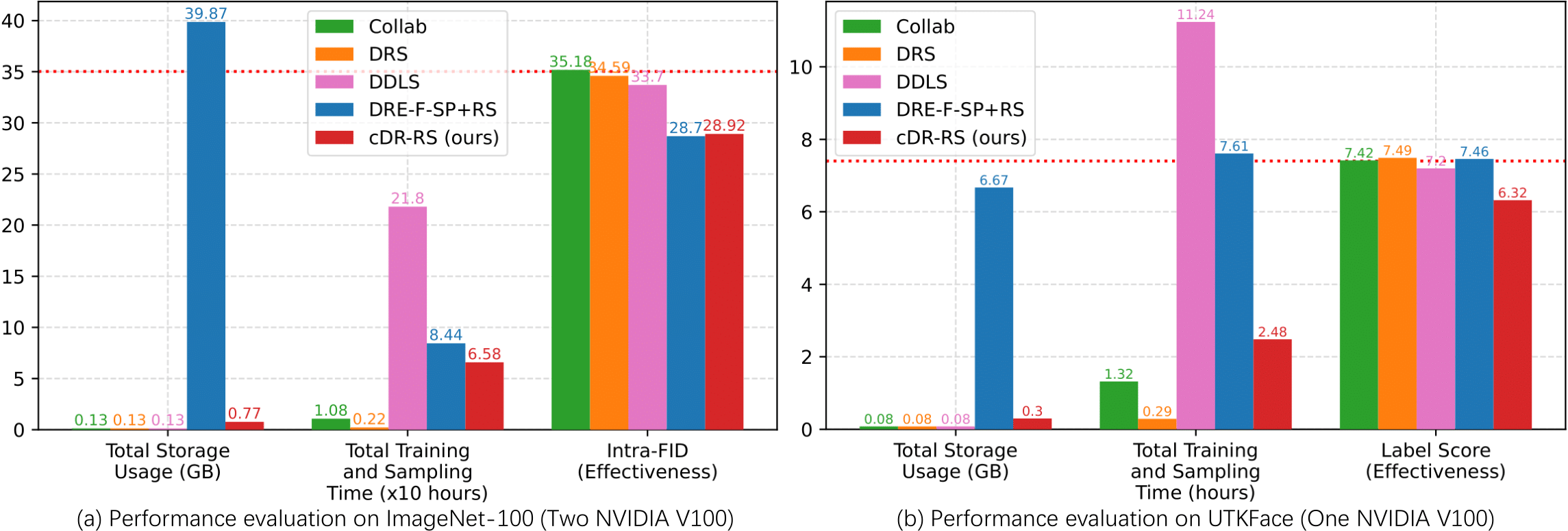}
	\caption{\textbf{The efficiency and effectiveness comparisons between the proposed cDR-RS and four baseline methods in sampling 100,000 and 60,000 fake images from BigGAN \cite{brock2018large} (a) and CcGAN \cite{ding2021ccgan, ding2020continuous} (b), respectively.} The dashed red lines denote Intra-FID \cite{miyato2018cgans} and Label Score \cite{ding2021ccgan, ding2020continuous} of sampling from cGANs without subsampling or refining in (a) and (b), respectively. We prefer lower Intra-FID and Label Score. Label Score is a metric to evaluate the discrepancy between the actual and conditioning labels of fake images. cDR-RS achieves state-of-the-art performances in sampling from cGANs while requiring only reasonable computational resources.}
	\label{fig:cover_fig}
\end{figure}

\textit{Generative adversarial networks} (GANs) \cite{goodfellow2014generative, li2021theoretical,chen2020adversarial,faezi2021degan} are a popular type of generative model for image synthesis, aiming to estimate the marginal distribution of images. As an extension and an essential family of GANs, \textit{conditional GANs} (cGANs) \cite{mirza2014conditional} intend to estimate the image distribution given some conditions. \rev{\textbf{In this paper, we focus on two main types of cGANs, i.e., \textit{class-conditional GANs} \cite{odena2017conditional, miyato2018cgans, brock2018large, zhang2019self, ali2019mfc,xu2021conditional} and \textit{continuous conditional GANs} (CcGANs) \cite{ding2021ccgan, ding2020continuous}.}} Class-conditional GANs such as BigGAN \cite{brock2018large} take class labels (i.e., categorical variables) as conditions. CcGANs \cite{ding2021ccgan, ding2020continuous} are a new type of cGANs, which take continuous, scalar variables (termed regression labels) as conditions. Unconditional GANs and cGANs have many applications, such as single image super-resolution \cite{ledig2017photo,dharejo2021twist}, image translation \cite{isola2017image, lai2021unsupervised, romero2019smit}, text-to-image synthesis \cite{ruan2021dae,zhou2021survey}, and 3D shape generation \cite{nobari2022range, triess2022point}.  

Recent advances in unconditional GANs \cite{karras2019style, karras2020analyzing} and class-conditional GANs \cite{brock2018large, ding2021ccgan, ding2020continuous} generally enable these models to generate high-quality images. Nevertheless, low-quality images still appear frequently even with such advanced GAN models during image generation. We would like to emphasize that the image quality discussed in this paper are three-fold \cite{miyato2018cgans, zhang2019self, devries2019evaluation, ding2021ccgan, ding2020continuous}: \textbf{(1) visual quality}, \textbf{(2) diversity}, and \textbf{(3) label consistency}. Label consistency applies to cGANs only and is defined as the consistency of generated images with respect to the conditioning label. Notably, CcGANs \cite{ding2021ccgan, ding2020continuous} often suffer from low label consistency because they are designed to sacrifice label consistency for better visual quality and higher diversity. Ding et al. \cite{ding2021ccgan, ding2020continuous} proposes Label Score to measure label consistency for CcGANs, where a lower Label Score implies higher label consistency and vice versa. 

To enhance the image quality of unconditional GANs, increasing attention has been paid to improve the sampling strategy of pre-trained unconditional GANs via \textbf{subsampling} or \textbf{refining}. Refining methods (e.g., Collab \cite{liu2020collaborative}) aim to refine visually unrealistic fake images to improve visual quality. For example, Collab \cite{liu2020collaborative} refines an intermediate hidden map of a generator to remove artifacts in generated images by using information from a trained discriminator. Subsampling methods (e.g., DRS \cite{azadi2018discriminator}, DRE-F-SP+RS \cite{ding2020subsampling}, DDLS \cite{che2020your}) remove visually unrealistic images to improve visual quality and adjust the likelihood of generated images to increase diversity. To accomplish subsampling, \textit{discriminator rejection sampling} (DRS) \cite{azadi2018discriminator} accepts or rejects a fake image by \textit{rejection sampling} (RS). DRS requires an accurate \textit{density ratio estimation} (DRE), however, because the DRE step in DRS relies on the assumption of optimality of the discriminator. Thus DRS may not perform well if the discriminator is far from optimal. DRS is also inapplicable to some GANs such as MMD-GAN \cite{li2017mmd}. Ding et al. improves DRS by proposing \textit{density ratio estimation in the feature space with Softplus loss} (DRE-F-SP) \cite{ding2020subsampling}, which does not require an optimal discriminator and is applicable to various GANs. Then, with RS as the sampler, \cite{ding2020subsampling} introduces DRE-F-SP+RS to subsample GANs. Besides these density ratio-based methods, \textit{discriminator driven latent sampling} (DDLS) \cite{che2020your} proposes to accept or reject samples via an energy-based model defined in the latent space of the generator in a GAN. The methods discussed above, however, are not designed for class-conditional GANs or CcGANs.

An intuitive approach to better sample from class-conditional GANs or CcGANs is applying unconditional methods described above for each distinct class or regression label. Unfortunately, this approach may be inefficient or even impractical, especially when many distinct classes or regression labels exist. For example, if we apply DRE-F-SP+RS \cite{ding2020subsampling} to sample from BigGAN \cite{brock2018large} trained on ImageNet-100 \cite{cao2017hashnet} (a subset of ImageNet \cite{imagenet_cvpr09} with 100 classes), we need to fit 100 density ratio models separately. As visualized in \cref{fig:cover_fig}(a), this is often time-consuming (84.4 hours) and it requires a large storage space ($39.97$ GB). Another noticeable example is DDLS \cite{che2020your}, which spends 218 hours subsampling BigGAN trained on ImageNet-100. Furthermore, as shown in \cref{fig:cover_fig}(b) and \Cref{sec:experiment_regression}, the unconditional approach is usually ineffective in sampling from CcGANs because it is not designed to solve the label inconsistency problem suffered by CcGANs. Moreover, although DRE-F-SP+RS may perform well in subsampling class-conditional GANs (e.g., \cref{fig:cover_fig}(a)), it is not suitable for subsampling CcGANs for two reasons: it lacks a suitable feature extraction method for regression datasets, and it cannot sample from CcGANs conditional on labels that are unseen in the training phase. The ineffectiveness or inefficiency of applying unconditional sampling methods to class-conditional GANs or CcGANs is demonstrated in our empirical study in Section \ref{sec:experiment}.

Recently, \cite{mo2019mining} proposes a rejection sampling scheme, called GOLD, to subsample the \textit{auxiliary classifier GAN} (ACGAN) \cite{odena2017conditional}, a special class-conditional GAN. GOLD is based on the gap in log-densities that measures the discrepancy between the actual image distribution and the fake image distribution of given samples. \cite{mo2019mining} shows that, empirically, GOLD can improve the performance of ACGAN in the class-conditional image synthesis. However, this method does not apply to the most recent cGANs (e.g., BigGAN \cite{brock2018large} and CcGAN \cite{ding2021ccgan, ding2020continuous}) for three reasons. First, its purpose, algorithm, and official codes are only designed for ACGAN. Second, GOLD relies on a strong assumption that the optimal discriminator is $D^*(\bm{x})=p_r(\bm{x})/(p_r(\bm{x})+p_g(\bm{x}))$ (Sec. 3.1 of \cite{mo2019mining}). Unfortunately, this assumption holds only if the discriminator is trained with the vanilla loss (Eq. (1) in \cite{goodfellow2014generative}) \cite{azadi2018discriminator, goodfellow2014generative}. It is invalid with GANs trained with other losses, e.g., hinge loss, Wasserstein loss, MMD loss, etc. Third, GOLD is inappropriate for subsampling CcGANs, because CcGANs do not have an ACGAN architecture, and GOLD does not provide a mechanism to control the label inconsistency problem suffered by CcGANs.

To improve the overall image quality of class-conditional GANs and CcGANs effectively and efficiently, we propose the \textit{conditional density ratio-guided rejection sampling} (cDR-RS). Our contributions can be summarized as follows:
\begin{itemize}
	\item In \Cref{sec:cDRE-F-cSP}, we propose a DRE scheme to estimate an image's density ratio conditional on a class or regression label. We first introduce a new feature extraction method for images with regression labels, when the feature extraction mechanism in DRE-F-SP \cite{ding2020subsampling} is not applicable. Then, we propose a novel \textit{conditional Softplus} (cSP) loss, which enables us to estimate density ratios conditional on different class/regression labels by fitting only one density ratio model. This density ratio model takes as input both the high-level features and the class/regression label of an image and outputs the density ratio conditional on the given label. 
	
	\item To analyze the proposed cSP loss theoretically, we derive in \Cref{sec:cDRE_error_bound} the error bound of a density ratio model trained with the proposed cSP loss.
	
	\item In \Cref{sec:cDR-RS}, we propose a novel rejection sampling scheme to subsample class-conditional GANs and CcGANs. A filtering scheme is also proposed in \Cref{sec:filtering_CcGAN} for CcGANs to increase fake images' label consistency without sacrificing diversity. By only tuning one hyper-parameter of this filtering scheme, we can easily control the trade-off between label consistency and diversity according to users' needs.
	
	\item In \Cref{sec:experiment}, extensive experiments on five benchmark datasets and different GAN architectures convincingly demonstrate the state-of-the-art performances of the proposed subsampling scheme over baseline methods. 
\end{itemize}

\section{Related works}\label{sec:related_work}

\subsection{Conditional generative adversarial networks}\label{sec:related_GANs}


\textit{Conditional GANs} (cGANs), first proposed in \cite{mirza2014conditional}, extend GANs \cite{goodfellow2014generative} to the conditional image synthesis setting, where a condition $y$ is fed into both the generator and discriminator networks. Mathematically, cGANs aim to estimate the density function $p_r(\bm{x}|y)$ of the actual conditional image distribution. The estimated density function $p_g(\bm{x}|y)$ is the density of the fake conditional image distribution induced by the generator network. The conditional $y$ is often a categorical variable such as a class label, and cGANs with class labels as conditions are also known as \textit{class-conditional GANs}. Class-conditional GANs have been widely studied in the literature \cite{odena2017conditional, miyato2018cgans, brock2018large, zhang2019self}. State-of-the-art class conditional GANs (e.g., BigGAN \cite{brock2018large}) can generate photo-realistic images for a given class. However, GANs conditional on regression labels have been rarely studied due to two problems. First, very few (even zero) real images exist for some regression labels. Second, since regression labels are continuous and infinitely many, they cannot be embedded by one-hot encoding like class labels. To solve these two problems, \cite{ding2021ccgan, ding2020continuous} propose the CcGAN framework, which introduces novel empirical cGAN losses and label input mechanisms. The novel empirical cGAN losses, consisting of the \textit{hard vicinal discriminator loss} (HVDL), the \textit{soft vicinal discriminator loss} (SVDL), and a new generator loss, are developed to solve the first problem. The second problem is solved by a \textit{naive label input} (NLI) mechanism and an \textit{improved label input} (ILI) mechanism. The effectiveness of CcGAN has been demonstrated on diverse datasets. Class-conditional GANs \cite{odena2017conditional, miyato2018cgans, brock2018large, zhang2019self} and CcGANs \cite{ding2021ccgan, ding2020continuous}, as two main types of cGANs, are our focus in this paper.


\subsection{Subsampling GANs}\label{sec:related_DRE-F-SP}

Among existing sampling methods for unconditional GANs, DRE-F-SP+RS proposed by \cite{ding2020subsampling} can achieve state-of-the-art sampling performance. This subsampling framework consists of two components: a density ratio estimation method, termed DRE-F-SP, and a rejection sampling scheme. DRE-F-SP aims to estimate the density ratio function $r^*(\bm{x}):=p_r(\bm{x})/p_g(\bm{x})$ based on $N^r$ real images $\bm{x}^r_1,\bm{x}^r_2,\dots,\bm{x}^r_{N^r}\sim p_r(\bm{x})$ and $N^g$ fake images $\bm{x}^g_1,\bm{x}^g_2,\dots,\bm{x}^g_{N^g}\sim p_g(\bm{x})$, where $p_r(\bm{x})$ and $p_g(\bm{x})$ are the density functions of the actual and fake conditional image distributions, respectively. Based on the estimated density ratios, to push $p_g$ towards $p_r$, rejection sampling (RS) is used to sample from the trained GAN model. Empirical studies in \cite{ding2020subsampling} show that DRE-F-SP+RS substantially outperforms existing sampling methods (e.g., DRS \cite{azadi2018discriminator}) in subsampling different types of unconditional GANs. 

As the key component of DRE-F-SP+RS, DRE-F-SP first trains a specially designed feature extractor $\phi$ on a set of real images with class labels under the cross-entropy loss. As visualized in \cref{fig:mod_resnet34}), $\phi$ is a \textit{convolutional neural network} (CNN) adapted from ResNet-34 \cite{he2016deep}. The network architecture of this feature extractor is adjusted so that the dimension of one hidden map $\bm{h}$ equals that of the input image $\bm{x}$, thus ensuring that the Jacobian determinant $\det({\partial \bm{h}}/{\partial \bm{x}})$ exists. The feature extractor defines a mapping of an image $\bm{x}$ to a high-level feature $\bm{h}$, i.e., $\bm{h} = \phi(\bm{x})$, where $\phi$ is assumed invertible and the absolute value of the Jacobian determinant $|\det({\partial \bm{h}}/{\partial \bm{x}})|$ is assumed positive. Then, DRE-F-SP estimates the density ratio of an image in the feature space defined by $\phi$ rather than in the pixel space. Specifically, DRE-F-SP models the actual density ratio function in the feature space by a 5-layer multilayer perceptron (MLP-5). \cite{ding2020subsampling} also proposes a novel loss function called Softplus (SP) loss to train this MLP-5. Finally, compositing $\phi(\bm{x})$ and MLP-5 leads to an estimate of $r^*(\bm{x})$.

Though DRE-F-SP+RS has been demonstrated both effective and efficient in subsampling unconditional GANs \cite{ding2020subsampling}; as discussed in \Cref{sec:intro}, however, several crucial challenges prevent us from applying it effectively to class-conditional GANs and CcGANs:
\begin{enumerate}[(C1)]
	\item DRE-F-SP+RS becomes very inefficient if many distinct class/regression labels exist;
	\item The feature extractor in DRE-F-SP+RS is inapplicable to images with regression labels only;
	\item DRE-F-SP+RS does not have a mechanism to control the label inconsistency problem suffered by CcGANs;
	\item DRE-F-SP+RS cannot sample from CcGANs conditional on regression labels that are unseen in the training set.
\end{enumerate}

These challenges are addressed by the proposed method, described next.

\section{Proposed method}\label{sec:method}
As discussed in Section \ref{sec:intro}, due to their ineffectiveness or inefficiency, the unconditional subsampling or refining methods (e.g., DRS \cite{azadi2018discriminator}, Collab \cite{liu2020collaborative}, DDLS \cite{che2020your} and DRE-F-SP+RS \cite{ding2020subsampling}) may be impractical for sampling from class-conditional GANs and CcGANs. Moreover, the only existing conditional subsampling method \cite{mo2019mining} is designed for ACGAN \cite{odena2017conditional} and cannot be applied to other cGANs. Motivated by these limitations, in this section, we propose an effective and efficient dual-functional subsampling method, which is suitable for both class-conditional GANs and CcGANs regardless of the GAN architecture and the number of distinct class or regression labels. 

To overcome the four crucial challenges listed at the end of \Cref{sec:related_DRE-F-SP}, the proposed method has to go well beyond straightforward extension of \cite{ding2020subsampling} to class-conditional GANs and CcGANs.  In this article we propose: 1) a novel loss for conditional DRE (for solving (C1) and (C4)); 2) a new feature extractor for CcGANs (for solving (C2)); and 3) a novel filtering scheme to control label inconsistency (for solving (C3)). Consequently, compared with \cite{ding2020subsampling}, the proposed novel methodology is substantially more efficient and is able to subsample CcGANs.

To aid the reader, we summarize in \Cref{tab:definition_symbols} the essential notation used in this paper along with definitions.

\begin{table}[!h]
	\centering
	\caption{\textbf{Notation and definitions.}}
	\begin{adjustbox}{width=1\textwidth}
		\begin{tabular}{c|l}
			\hline\hline
			Notation & Definitions \\
			\hline
			$\bm{x}$ & an image at $C\times H\times W$ resolution, which may have a subscript or a superscript, e.g., $\bm{x}_i^r$ for real image $i$  \\
			\hline
			$\hat{\bm{x}}$ & a reconstructed image at $C\times H\times W$ resolution, which may have a subscript or a superscript, e.g., $\bm{x}_i^r$ \\
			\hline
			$\phi(\bm{x})$   &  the feature extractor, i.e., a pre-trained, invertible neural network \\
			\hline
			$\bm{h}$ &  \begin{tabular}[l]{@{}l@{}} the high-level feature (a $C\times H\times W$ by 1 vector) extracted by $\phi(\bm{x})$ from an image, which may have \\ a subscript or a superscript, e.g., $\bm{h}_i^r$\end{tabular} \\
			\hline
			$y$ & a class/regression label, which may have a subscript or a superscript, e.g., $y_i^r$  \\
			\hline
			$\hat{y}$ & a predicted class/regression label, which may have a subscript or a superscript, e.g., $\hat{y}_i^r$  \\
			\hline
			$p_r(\bm{x}|y)$ & the density function of the real images conditional on $y$  \\
			\hline
			$p_g(\bm{x}|y)$ & the density function of the fake images conditional on $y$  \\
			\hline
			$q_r(\bm{h}|y)$ & the density function of the real features conditional on $y$  \\
			\hline
			$q_g(\bm{h}|y)$ & the density function of the fake features conditional on $y$  \\
			\hline
			$p(y)$ & the density function of real/fake labels  \\
			\hline
			$q_r(\bm{h},y)$ & the joint density function of the real features and labels  \\
			\hline
			$q_g(\bm{h},y)$ & the joint density function of the fake features and labels  \\
			\hline
			$\det\left(\partial\bm{h} / \partial\bm{x}\right)$ & the determinant of the Jacobian matrix $\partial\bm{h} / \partial\bm{x}$ \\
			\hline
			$\sigma(t)$ &  Sigmoid function, i.e., $\sigma(t)={e^t}/{1+e^t}$   \\
			\hline
			$\eta(t)$   &  Softplus function, i.e., $\eta(t)=\ln{(1+e^t)}$      \\
			\hline
			$r^*(\bm{x}|y)$   &   the actual conditional density ratio function  of images  \\
			\hline
			$\psi^*(\bm{h}|y)$   &   the actual conditional density ratio function of extracted features  \\
			\hline
			$\psi(\bm{h}|y)$   &   conditional density ratio model (neural network)  \\
			\hline
			$\mathcal{L}_c(\psi)$ & the proposed conditional Softplus (cSP) loss \\
			\hline
			$\widehat{\mathcal{L}}_c(\psi)$ & the empirical version of the cSP loss \\
			\hline
			$\Psi$   &  a set of functions that can be represented by $\psi(\bm{h}|y)$  \\
			\hline
			$\tilde{\psi}$ & $\tilde{\psi}=\arg\min_{\psi\in\Psi}\mathcal{L}_c(\psi)$ \\
			\hline
			$\hat{\psi}$ & $\hat{\psi}=\arg\min_{\psi\in\Psi}\widehat{\mathcal{L}}_c(\psi)$ \\
			\hline
			$\widehat{Q}_c(\psi)$   &  the penalty term used when training $\psi$    \\
			\hline
			$\lambda$ & a positive hyper-parameter to control the penalty strength of $\widehat{Q}_c(\psi)$  \\
			\hline
			$\lambda^\prime$ & a positive hyper-parameter to control the sparsity of extracted features  \\
			\hline
			$\mathcal{Y}_y^{\zeta}$ & $\mathcal{Y}_y^{\zeta}:=[y-\zeta, y+\zeta]$, i.e., the neighborhood of a regression label $y$ with width $\zeta$ \\
			\hline
			$\zeta$ & a positive hyper-parameter to control the width of $y$'s neighborhood $\mathcal{Y}_y^{\zeta}$ \\
			\bottomrule
		\end{tabular}%
	\end{adjustbox}
	\label{tab:definition_symbols}%
\end{table}%

\subsection{Conditional density ratio estimation in feature space with conditional Softplus loss} \label{sec:cDRE-F-cSP}
In this section, we introduce cDRE-F-cSP, a novel \textit{conditional density ratio estimation} (cDRE) method. Assume we have $N^r$ real image-label pairs $(\bm{x}_1^r, y_1^r),  \dots, (\bm{x}_{N^r}^r, y_{N^r}^r)$ and $N^g$ fake image-label pairs $(\bm{x}_1^g, y_1^g), \dots, (\bm{x}_{N^g}^g, y_{N^g}^g)$, where $\bm{x}_i^r$ and $\bm{x}_i^g$ can be seen as samples drawn from $p_r(\bm{x}|y_i^r)$ and $p_g(\bm{x}|y_i^g)$, respectively. Based on these observed samples, we aim to estimate
\begin{equation}
    \label{eq:true_density_ratio_function}
    r^*(\bm{x}|y):=\frac{p_r(\bm{x}|y)}{p_g(\bm{x}|y)}.
\end{equation}

\begin{figure}[t]
	\centering
	\includegraphics[width=0.6\linewidth]{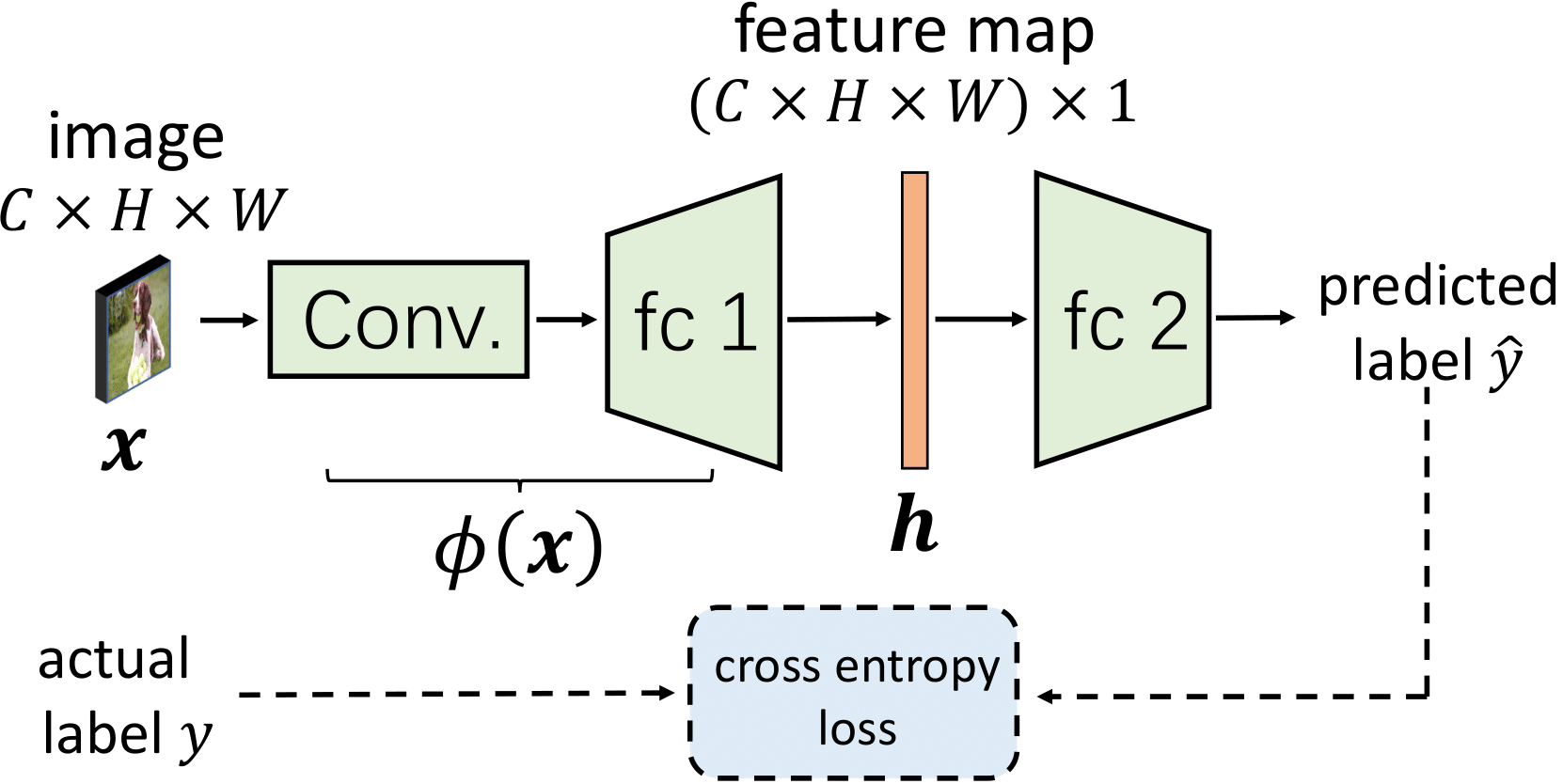}
	\caption{\textbf{The network architecture of the feature extractor $\phi(\bm{x})$ for subsampling class-conditional GANs.} The feature extractor $\phi(\bm{x})$ consists of a convolutional block (Conv.) and a fully-connected block (fc 1), where Conv. is the convolutional block of ResNet-34 and fc 1 increases the dimension of features to $(C\times H\times W)\times 1$ for equal dimensionality. An extra fully-connected block (fc 2) predicts the class label $y$ of the input image $\bm{x}$ in terms of the extracted feature $\bm{h}$. The whole model ($\phi$ and fc 2) is trained under the cross entropy loss. Although the invertibility of $\phi$ is a necessary assumption for Eq.\ \eqref{eq:cdre_equivalence} to hold, and the vanilla ResNet-34 architecture is not invertible, \cite{ding2020subsampling} shows that $\phi(\bm{x})$ adapted from ResNet-34 still works very well as the feature extractor in subsampling unconditional GANs. One possible reason is that the adjustment of the ResNet-34 architecture to ensure equal dimensionality improves the invertibility of $\phi$. Therefore, we also employ such a modified ResNet-34 in subsampling class-conditional GANs.}
	\label{fig:mod_resnet34}
\end{figure}

\begin{figure}[t]
	\centering
	\includegraphics[width=0.53\linewidth]{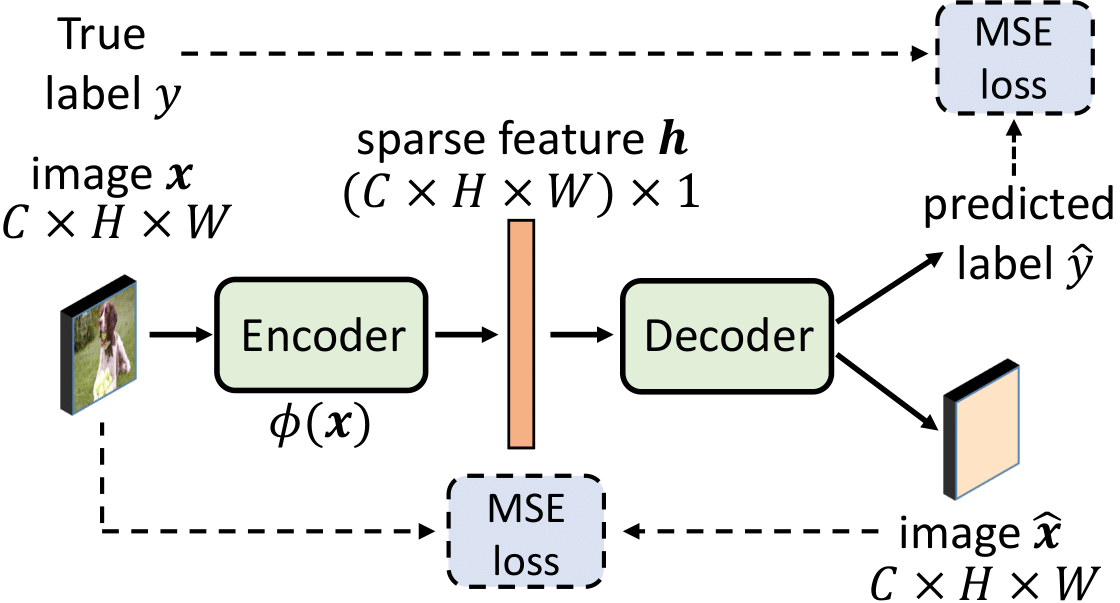}
	\caption{\textbf{The network architecture of the sparse autoencoder (SAE) for feature extraction in subsampling CcGANs.} Usually, the bottleneck dimension of SAE is much smaller than the input image, but the validity of Eq. \eqref{eq:cdre_equivalence} requires they have equal dimensions. To prevent SAE from overfitting training data, we penalize the mean of the features to encourage sparsity. With ReLU as the last layer, the features $\bm{h}=[h_1,...h_m]$ extracted by the encoder are non-negative. Thus, their mean  $(\sum_i^mh_i)/m$ is $(\sum_i^m|h_i|)/m$, i.e., the $L_1$ penalty on $\bm{h}$ commonly used to encourage sparsity. Additionally, this SAE also has an extra branch to predict the regression label of $\bm{x}$. Both the sparsity penalty and the extra prediction branch are used to avoid overfitting. Note that, $\phi(\bm{x})$ is trained to be invertible and Decoder can be seen as its inverse function.}
	\label{fig:sparseAE}
\end{figure}

Like DRE-F-SP \cite{ding2020subsampling}, we conduct cDRE in a feature space learned by a \textit{pre-trained}, \textit{invertible} CNN $\phi$. DRE-F-SP \cite{ding2020subsampling} trains a classification CNN adapted from ResNet-34 on some real images with class labels to extract high-level features for DRE (refer to \Cref{sec:related_DRE-F-SP} and \cref{fig:mod_resnet34}). This mechanism also applies to class-conditional GANs; however, it is inapplicable to CcGANs since regression datasets may not have class labels. Therefore, when subsampling CcGANs, we develop a specially designed \textit{sparse autoencoder} (SAE) to extract features whose architecture is visualized in \cref{fig:sparseAE}. The encoder with ReLU \cite{glorot2011deep} as the final layer is treated as $\phi$ to extract sparse high-level features from images. The bottleneck dimension of the sparse autoencoder equals the dimension of the flattened input image. The decoding process is trained to reconstruct the input image and predict the regression label of the input image. The training loss of SAE is the summation of three loss components: 
\begin{enumerate}[(1)]
    \item The \textit{mean square error} (MSE) between the input image $\bm{x}$ and the reconstructed image $\hat{\bm{x}}$, i.e., $\frac{1}{C\cdot H\cdot W}\|\bm{x}-\hat{\bm{x}}\|^2_2$, where $\|\cdot\|_2$ is the Euclidean norm;
    
    \item The squared error between the actual regression label $y$ and the predicted regression label $\hat{y}$, i.e., $(y-\hat{y})^2$;
    
    \item The product of a positive constant $\lambda^\prime$ and the $L_1$ norm of $\bm{h}$, i.e., $\lambda^\prime\cdot\frac{1}{C\cdot H\cdot W}\|\bm{h}\|_1$, where $\|\cdot\|_1$ is the $L_1$ norm. The constant $\lambda^\prime$ controls the sparsity and $\lambda^\prime=10^{-3}$ often empirically works well.
    
\end{enumerate}
Specifically, we minimize the following training loss $\widehat{\mathcal{L}}_{SAE}$ in practice defined based on $N^r$ real image-label pairs:
\begin{equation}
    \widehat{\mathcal{L}}_{SAE} = \frac{1}{N^r\cdot C\cdot H\cdot W}\sum_{i=1}^{N^r}\|\bm{x}_i^r-\hat{\bm{x}}_i^r\|_2^2 + \frac{1}{N^r}\sum_{i=1}^{N^r}(y_i^r-\hat{y}_i^r)^2 + \lambda^\prime \cdot \frac{1}{N^r\cdot C\cdot H\cdot W}\sum_{i=1}^{N^r}\|\bm{h}^r_i\|_1,
\end{equation}
where $\bm{x}_i^r$ is the $i$-th real image, $\hat{\bm{x}}_i^r$ is the $i$-th reconstructed real image, $y_i^r$ is the $i$-th real label, $\hat{y}^r_i$ is the predicted label of $\bm{x}_i^r$, and $\bm{h}^r_i$ is the extracted feature vector of $\bm{x}_i^r$. Usually, the bottleneck dimension of an autoencoder is much smaller than the input image, but DRE in feature space \cite{ding2020subsampling} requires they have equal dimensions. To prevent the proposed SAE from overfitting the training data, we design an extra branch to predict regression labels and an $L_1$ regularizer to encourage the sparsity of extracted features. The additional predictor is also applied in the filtering scheme proposed in \Cref{sec:filtering_CcGAN}.

Next, we propose a formulation of the \textbf{actual} conditional density ratio function in the feature space: Given that (1) $\bm{h}$ and the flattened $\bm{x}$ have equal dimension, (2) the feature extractor $\phi$ is invertible, and (3) the Jacobian determinant $|\det({\partial \bm{h}}/{\partial \bm{x}})|$ is nonzero, then
\begin{equation}
	\label{eq:cdre_equivalence}
	\scalemath{1}{
		\begin{aligned}
			\psi^*(\bm{h}|y)&:=\frac{q_r(\bm{h}|y)}{q_g(\bm{h}|y)}=\frac{q_r(\bm{h}|y)\cdot\left|\det\left(\frac{\partial\bm{h}}{\partial\bm{x}}\right)\right|}{q_g(\bm{h}|y)\cdot \left|\det\left(\frac{\partial\bm{h}}{\partial\bm{x}}\right)\right|} =\frac{p_r(\bm{x}|y)}{p_g(\bm{x}|y)} = r^*(\bm{x}|y),
		\end{aligned}
	}
\end{equation}
where $q_r(\bm{h}|y)$ and $q_g(\bm{h}|y)$ are respectively the density functions of the real and fake conditional distributions of high-level features. Based on \cref{eq:cdre_equivalence} and the pre-trained neural network $\phi(\bm{x})$, we model the conditional density ratio function $\psi^*(\bm{h}|y)$ in the feature space by a 5-layer MLP (MLP-5) denoted by $\psi(\bm{h}|y)$ with both $\bm{h}$ and its label $y$ as input. To train $\psi(\bm{h}|y)$, we propose the \textit{conditional Softplus} (cSP) loss: 
\begin{align}
	\mathcal{L}_c(\psi) = \mathbb{E}_{(\bm{h},y)\sim q_g(\bm{h},y)}\left[\sigma(\psi(\bm{h}|y)) \psi(\bm{h}|y) - \eta(\psi(\bm{h}|y)) \right]  - \mathbb{E}_{(\bm{h},y)\sim q_r(\bm{h},y)}\left[ \sigma(\psi(\bm{h}|y)) \right],
	\label{eq:SP_loss_cond}
\end{align}
where $\sigma$ and $\eta$ are Sigmoid and Softplus functions respectively, $q_g(\bm{h},y)=q_g(\bm{h}|y)p(y)$, $q_r(\bm{h},y)=q_r(\bm{h}|y)p(y)$, and $p(y)$ is the distribution of labels. This new loss extends the SP loss \cite{ding2020subsampling} to the conditional setting. The empirical approximation to \cref{eq:SP_loss_cond} is 
\begin{equation}
	\widehat{\mathcal{L}}_c(\psi) = \frac{1}{N_g}\sum_{i=1}^{N_g}\left[ \sigma(\psi(\bm{h}_i^g|y_i^g))\psi(\bm{h}_i^g|y_i^g) -\eta(\psi(\bm{h}_i^g|y_i^g)) \right]  - \frac{1}{N_r}\sum_{i=1}^{N_r}\sigma(\psi(\bm{h}_i^r|y_i^r)).
	\label{eq:emp_SP_loss_cond}
\end{equation}
Similar to DRE-F-SP, to prevent $\psi(\bm{h}|y)$ from overfitting the training data, a natural constraint applied to $\psi(\bm{h}|y)$ is 
\begin{equation}
	\label{eq:cond_constraint}
	\mathbb{E}_{\bm{h}\sim q_g(\bm{h}|y)} [\psi(\bm{h}|y)] =1.
\end{equation}
If \cref{eq:cond_constraint} holds, then
\begin{equation}
	\label{eq:joint_constraint}
	\mathbb{E}_{(\bm{h},y)\sim q_g(\bm{h},y)} [\psi(\bm{h}|y)] = 1.
\end{equation}
An empirical approximation to \cref{eq:joint_constraint} is
\begin{equation}
	\label{eq:emp_joint_constraint}
	\frac{1}{N^g}\sum_{i=1}^{N^g} \psi(\bm{h}_i^g|y_i^g) = 1.
\end{equation}
Therefore, in practice, we minimize the penalized version of \cref{eq:emp_SP_loss_cond} as follows:
\begin{equation}
	\label{eq:penalized_SP_loss_cond}
	\min_{\psi}\left\{ \widehat{\mathcal{L}}_c(\psi) + \lambda\widehat{Q}_c(\psi) \right\},
\end{equation}
where 
\begin{equation}
	\label{eq:penalty_Q_cond}
	\scalemath{1}{
		\widehat{Q}_c(\psi) = \left(\frac{1}{N^g}\sum_{i=1}^{N^g} \psi(\bm{h}_i^g|y_i^g) - 1 \right)^2.
	}
\end{equation}

An algorithm shown in Alg.\ \ref{alg:cDRE-F-cSP} is used to implement cDRE-F-cSP in practice. Both \cite{ding2020subsampling} and our empirical study show that the fitted (conditional) density ratio model often performs well in subsampling if $\lambda\in[0,0.1]$. In our experiments, $\lambda$ is set as $10^{-2}$ or $10^{-3}$ empirically. 

\begin{remark}
	Although the invertibility of the feature extractor $\phi$ is a necessary assumption for Eq.\ \eqref{eq:cdre_equivalence} to hold, an empirical study in \cite{ding2020subsampling} shows that the modified ResNet-34 (refer to \Cref{sec:related_DRE-F-SP} and \cref{fig:mod_resnet34}) still works very well as the feature extractor in subsampling unconditional GANs. One possible reason is that the adjustment of the ResNet-34 architecture to ensure equal dimensionality improves the invertibility of the feature extractor. Therefore, we also employ such a modified ResNet-34 in subsampling class-conditional GANs. For CcGANs, we propose an SAE to extract features, where the encoder is trained to be invertible (i.e., the decoder can be seen as the inverse function of the encoder). A similar SAE may also be used in subsampling class-conditional GANs, but it seems unnecessary in practice and is not used in the experiments of \Cref{sec:experiment_classification}.
\end{remark}


\begin{remark}\label{rmk:how_to_input_condition}
    A condition $y$ can be incorporated into the conditional density ratio model $\psi(\bm{h}|y)$ through concatenation. Specifically, we first convert the class/regression label $y$ into an embedding vector. In the class-conditional scenario, $y$ is encoded as a one-hot vector. In the CcGAN scenario, the condition $y$ is a continuous scalar, so it cannot be one-hot encoded. In this case, we adopt the \textit{improved label input} (ILI) method in \cite{ding2020continuous} which projects the scalar $y$ into a high-dimensional embedding space. After the embedding, to input the condition into $\psi(\bm{h}|y)$, the encoded class/regression label is then concatenated with the input (i.e., the extracted feature vector $\boldsymbol{h}$) or a hidden map of $\psi(\bm{h}|y)$. Please refer to \Cref{tab:cifar10_cMLP5} in Appendix for an example of how to input $y$ into $\psi(\bm{h}|y)$.
\end{remark}

\begin{algorithm}[t] 
	\footnotesize
	\SetAlgoLined
	\KwData{$N^r$ real image-label pairs $\{(\bm{x}^r_1, y^r_1),\cdots,(\bm{x}^r_{N^r}, y^r_{N^r})\}$, a generator $G$, a pre-trained feature extractor $\phi(\bm{x})$, a 5-layer MLP $\psi(\bm{h}|y)$ and a preset hyperparameter $\lambda$.}
	\KwResult{A trained conditional density ratio model $r(\bm{x}|y)=\psi(\phi(\bm{x})|y)=\psi(\bm{h}|y)$. }
	Initialize $\psi$\;
	\For{$k=1$ \KwTo $K$}{
		Sample a mini-batch of $m$ \textbf{real} image-label pairs $\{(\bm{x}^r_{(1)},y^r_{(1)}),\cdots,(\bm{x}^r_{(m)},y^r_{(m)})\}$ from $\{(\bm{x}^r_1, y^r_1),\cdots,(\bm{x}^r_{N^r}, y^r_{N^r})\}$\;
		Sample a mini-batch of $m$ \textbf{fake} image-label pairs $\{(\bm{x}^g_{(1)},y^g_{(1)}),\cdots,(\bm{x}^g_{(m)},y^g_{(m)})\}$ from $G$\;
		Update $\psi$ via mini-batch SGD or a variant with the gradient of \cref{eq:penalized_SP_loss_cond}, i.e., $\bigtriangledown \left\{ \widehat{\mathcal{L}}_c(\psi) + \lambda\widehat{Q}_c(\psi) \right\}$.
	}
	\caption{Optimization algorithm for conditional density ratio model training in cDRE-F-cSP.}
	\label{alg:cDRE-F-cSP}
\end{algorithm}

\subsection{Error bound}\label{sec:cDRE_error_bound}
In this section, we derive the error bound of a density ratio model $\psi(\bm{h}|y)$ trained with the empirical cSP loss $\widehat{\mathcal{L}}_c(\psi)$. For simplicity, we ignore the penalty term in this analysis. 

Firstly, we introduce some notation. Let $\Psi=\{ \psi: \bm{h}\rightarrow \psi(\bm{h}|y) \}$ denote the hypothesis space of the density ratio model $\psi(\bm{h}|y)$. We also define $\tilde{\psi}$ and $\hat{\psi}$ as follows:
$\tilde{\psi}=\arg\min_{\psi\in\Psi}\mathcal{L}_c(\psi)$ and $ \hat{\psi}=\arg\min_{\psi\in\Psi}\widehat{\mathcal{L}}_c(\psi).$
Please note that the hypothesis space $\Psi$ may not cover the actual density ratio function $\psi^*$. Therefore, $\mathcal{L}_c(\tilde{\psi})-\mathcal{L}_c(\psi^*)\geq 0$. Denote by $\bm{\alpha}$ all learnable parameters of $\psi$ and assume $\bm{\alpha}$ is in a parameter space $\mathcal{A}$. Denote $\sigma(\psi(\bm{h}|y))\psi(\bm{h}|y)-\eta(\psi(\bm{h}|y))$ by $g(\bm{h}|y;\bm{\alpha})$. Let $\hat{\mathcal{R}}_{q_r(\bm{h},y),N^r}(\Psi)$ denote the empirical Rademacher complexity \cite{mohri2018foundations} of $\Psi$, which is defined based on independent feature-label pairs $\{ (\bm{h}_1^r,y_1^r), \dots, (\bm{h}_{N^r}^r,y_{N^r}^r) \}$ from $q_r(\bm{h},y)$. Then,
we derive the error bound of the conditional density ratio estimate $\hat{\psi}$ under \cref{eq:SP_loss_cond} as follows:
\begin{theorem}
	\label{thm:error_bound}
	If (i) $N^g$ is large enough, (ii) $\mathcal{A}$ is compact, (iii) $\forall g(\bm{h}|y; \bm{\alpha})$ is continuous at $\bm{\alpha}$, (iv) $\forall g(\bm{h}|y; \bm{\alpha})$, $\exists$ a function $g^u(\bm{h}|y)$ that does not depend on $\bm{\alpha}$, s.t. $|g(\bm{h}|y;\bm{\alpha})|\leq g^u(\bm{h}|y)$, and (iv) $\mathbb{E}_{(\bm{h},y)\sim q_g(\bm{h},y)}g^u(\bm{h}|y)<\infty$, then $\forall \delta\in(0,1)$ and $\forall \delta^\prime\in(0,\delta]$ with probability at least $1-\delta$,
    
    \begin{align}
		\mathcal{L}_c(\hat{\psi})-\mathcal{L}_c(\psi^*)  \leq \frac{1}{N^g} + \hat{\mathcal{R}}_{q_r(\bm{h},y),N^r}(\Psi) + 2\sqrt{\frac{4}{N^r}\log\left(\frac{2}{\delta^\prime}\right)} + \mathcal{L}_c(\tilde{\psi})-\mathcal{L}_c(\psi^*).
		\label{eq:error_bound}
	\end{align}
\end{theorem}

\begin{proof}
	The proof is in Supp.\ \ref{supp:proofs}.
\end{proof}

\begin{remark}
    \Cref{thm:error_bound} provides an error bound of a conditional density ratio model trained with the cSP loss.  $\hat{\mathcal{R}}_{q_r(\bm{h},y),N^r}(\Psi)$ on the right of \cref{eq:error_bound} implies that, a more complicated density ratio model has a more significant error bound, i.e., a worse generalization performance. Thus, we should not use an overly complicated density ratio model. It supports our proposed cDRE-F-cSP because we just need a small neural network (e.g., a shallow MLP) to model the density ratio function in the feature space.
\end{remark}

\subsection{cDR-RS: Conditional density ratio-guided rejection sampling for both class-conditional GANs and CcGANs}\label{sec:cDR-RS}

\begin{figure}[h]
	\centering
	\includegraphics[width=0.6\linewidth]{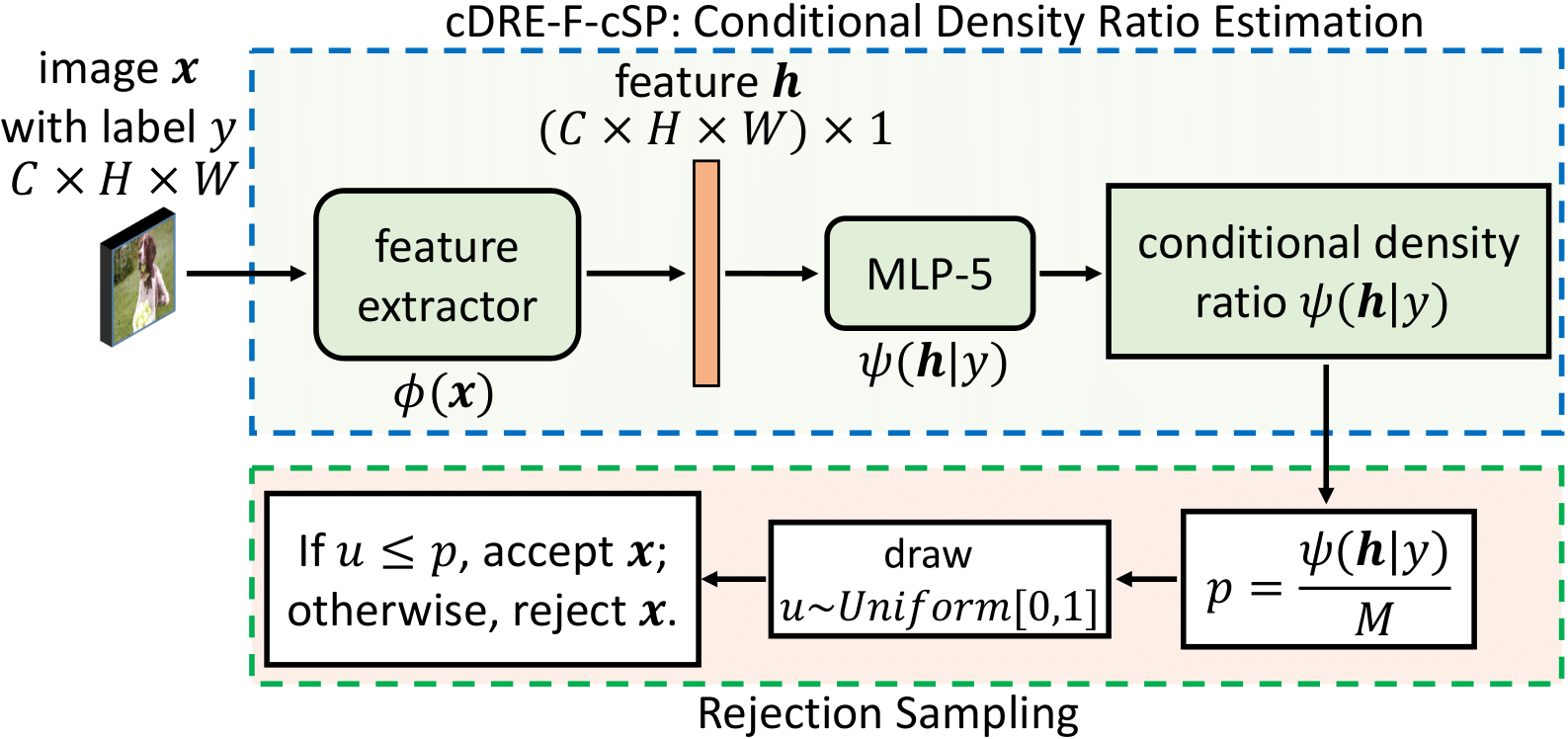}  
	\caption{\textbf{The workflow of cDR-RS has two sequential modules: cDRE-F-cSP and rejection sampling.} $M$ in the acceptance probability $p$ equals to $\max_{\bm{h}}\{q_r(\bm{h})/q_g(\bm{h})\}$, which can be estimated by evaluating $\psi(\bm{h}|y)$ on some burn-in samples before subsampling. We input the condition $y$ into $\psi(\bm{h}|y)$ via concatenation (please refer to \Cref{rmk:how_to_input_condition} and \Cref{tab:cifar10_cMLP5}). The rejection sampling scheme is conducted based on $\psi(\bm{h}|y)$ for each distinct label $y$ of interest.}
	\label{fig:workflow_cDR-RS}
\end{figure}

\rev{Based on the cDRE method proposed in Section \ref{sec:cDRE-F-cSP}, we develop a rejection sampling scheme, termed cDR-RS, to subsample class-conditional GANs and CcGANs. The workflow can be summarized in \cref{fig:workflow_cDR-RS} and Alg.\ \ref{alg:cDR-RS}. \textbf{This rejection sampling scheme is conducted based on $\psi(\bm{h}|y)$ for each distinct label $y$ of interest.} For example, in the ImageNet-100 \cite{cao2017hashnet} experiment, after fitting $\psi(\bm{h}|y)$ on real and fake image datasets, we conduct rejection sampling from the trained BigGAN within each class. Since the ImageNet-100 dataset consists of 100 classes, in total we will do 100 rejection sampling operations to get fake images, i.e., conducting one separate rejection sampling for each individual class. }



\begin{algorithm}[h]
	\footnotesize
	\SetAlgoLined
	\KwData{A generator $G$, a trained feature extractor $\phi(\bm{x})$, a trained conditional density ratio model $\psi(\bm{h}|y)$.}
	\KwResult{$images=\{N\text{ filtered fake images with label $y$}\}$}
	Generate $N^\prime$ burn-in fake images from $G$ conditional on label $y$.\;
	Estimate the density ratios of these $N^\prime$ fake images conditional on $y$ by evaluating $\psi(\phi(\bm{x})|y)$\;
	$M\leftarrow\max\{N^\prime \text{ estimated density ratios}\}$\;
	$images\leftarrow \emptyset$\;
	\While{$|images|<N$}{
		$\bm{x}\leftarrow \text{get a fake image with label $y$ from }G$\;
		$ratio\leftarrow \psi(\phi(\bm{x})|y)$\;
		$M\leftarrow\text{max}\{M,ratio\}$\;
		$p\leftarrow ratio/M$ (i.e., the acceptance probability in RS)\;
		$u\leftarrow \text{Uniform}[0,1]$\;
		\If{$u\leq p$}{
			$\text{Append}(\bm{x}, images)$\;
		}
	}
	\caption{Subsampling fake images with label $y$ by cDR-RS.}
	\label{alg:cDR-RS}
\end{algorithm}

\subsection{A filtering scheme to increase label consistency in subsampling CcGANs}\label{sec:filtering_CcGAN}
To solve the problem of insufficient data, CcGANs \cite{ding2021ccgan, ding2020continuous} use real images with labels in a vicinity of a conditioning regression label $y$ to estimate $p_r(\bm{x}|y)$. Consequently, the actual labels of some fake images sampled from $p_g(\bm{x}|y)$ may be far from $y$ (a.k.a label inconsistency). Unfortunately, the current subsampling scheme in \cref{fig:workflow_cDR-RS} may still accept these fake images if they have good visual quality or can contribute to the diversity increase (see \Cref{tab:effectiveness_analysis_regression}). An intuitive solution to this issue is to filter out these fake images with actual labels different from $y$, but this filtering may substantially decrease the diversity of fake images. 

\rev{Please note that the actual labels of fake images are unknown to us, therefore we need to estimate them in practice. Specifically, we first predict the labels of fake images via a pre-trained regression CNN (i.e., the composition of the encoder and predictor of SAE in \cref{fig:sparseAE}), and the predicted labels are assumed to be close to the actual labels. Then, we can estimate fake images' actual labels by their predicted labels.}

To increase label consistency without losing diversity, we propose a filtering scheme for cDR-RS, which is \rev{based on an assumption} that we may sample fake images with actual labels equal to $y$ from both $p_g(\bm{x}|y)$ and $p_g(\bm{x}|y^\prime)$ in the CcGAN sampling, where $y^\prime$ is close to $y$. To illustrate the proposed filtering scheme consider subsampling at $y$. We first replace $p_g(\bm{x}|y)$ with $p_g^{y, \zeta}(\bm{x})$ as the density of the proposal distribution, where $p_g^{y, \zeta}(\bm{x})$ stands for the density function of the distribution of fake images with predicted labels in $\mathcal{Y}_y^{\zeta}:=[y-\zeta, y+\zeta]$ and $\zeta$ is a hyper-parameter. Afterwards, the density ratio model $r(\bm{x}|y)$ is used to model $p_r(\bm{x}|y)/p_g^{y, \zeta}(\bm{x})$ and it is trained on fake images with predicted labels in $\mathcal{Y}_y^{\zeta}$. In the sampling phase, before conducting the rejection sampling process (\cref{fig:workflow_cDR-RS} and Alg.\ \ref{alg:cDR-RS}), we filter out fake images with predicted labels outside of $\mathcal{Y}_y^{\zeta}$. Please note that $\zeta$ controls the trade-off between label consistency and diversity. As shown in \cref{fig:UTKFace_effect_of_kappa}, a smaller $\zeta$ often leads to higher label consistency but lower diversity, while a larger $\zeta$ often leads to lower label consistency but higher diversity. We can adjust $\zeta$ according to our needs, but a good $\zeta$ should improve label consistency without decreasing diversity. In our experiment, $\zeta$ is set based on a vicinity parameter $m_{\kappa}$ of CcGANs \cite{ding2021ccgan, ding2020continuous}. Please refer to \Cref{rmk:rule_of_thumb_for_zeta} for a rule of thumb to select $\zeta$.

\begin{remark}[\textbf{A rule of thumb to select $\zeta$}]\label{rmk:rule_of_thumb_for_zeta}
    {\color{black}
    
    Let's first review the rule of thumb for selecting the vicinity hyper-parameter in CcGANs \cite{ding2021ccgan, ding2020continuous}. CcGANs estimate $p_r(\bm{x}|y)$ based on real images with labels in a vicinity of $y$. Let $\kappa$ denote the hyper-parameter to control the width of the hard vicinity of CcGANs. Denote by $N^r_{\text{uy}}$ the number of unique labels in the training set, and denote by $y^r_{[l]}$ the $l$-th smallest normalized distinct real label. Define $\kappa_{\text{base}}$ as $\kappa_{\text{base}}=\max\left(y^r_{[2]}-y^r_{[1]},  y^r_{[3]}-y^r_{[2]}, \dots,  y^r_{[N^r_{\text{uy}}]}-y^r_{[N^r_{\text{uy}}-1]} \right)$. Then, \cite{ding2021ccgan, ding2020continuous} set $\kappa$ as $\kappa=m_{\kappa}\kappa_{\text{base}}$, where $m_{\kappa}$ stands for 50\% of the minimum number of neighboring labels used for estimating $p_r(\bm{x}|y)$ given a label $y$. 
    
    As we discussed in \Cref{sec:filtering_CcGAN}, we estimate fake images' actual labels by their predicted labels. The predicted labels may deviate from the actual labels, but such deviation is assumed not significant. Although the prediction error may be small, it still exists and won't be zero. Therefore, when designing the proposal distribution in the rejection sampling, instead of using fake images with predicted labels exactly equal to $y$ (i.e., set $\zeta=0$), we consider fake images with predicted labels in a vicinity of $y$, i.e., $\mathcal{Y}_y^\zeta$. We expect that the actual labels of fake images in $\mathcal{Y}_y^\zeta$ should be equal or close to $y$. Furthermore, if we let $\zeta=0$, the training and sampling time may be very long. On UTKFace, we conduct an ablation study to show the effect of $\zeta$ on the training and sampling. Relevant results are summarized in \cref{fig:UTKFace_zeta_vs_total_time}. We can see a smaller $\zeta$ substantially increases the training and sampling time, which may make cDR-RS less efficient. Therefore, in practice, we usually use a relatively large $\zeta$, e.g., $3\times m_{\kappa}\kappa_{\text{base}}$. In other words, we let the vicinity $\mathcal{Y}_y^{\zeta}=[y-\zeta, y+\zeta]$ to be three times wider than the hard vicinity in CcGANs. Our experiments in \Cref{sec:experiment_regression} show that this rule of thumb can let cDR-RS effectively increase label consistency without losing diversity.
    
    }

    \begin{figure}[!htbp]
    	\centering
    	\includegraphics[width=0.5\textwidth]{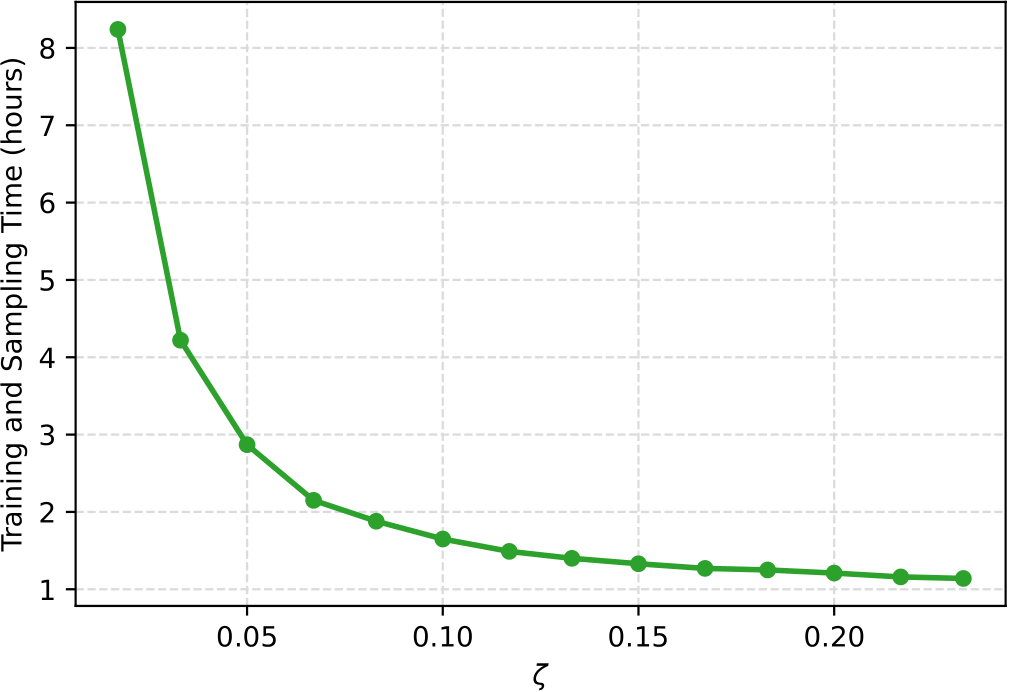}  
    	\caption{\textbf{The effect of $\zeta$ in the filtering scheme on the implementation time of cDR-RS on UTKFace.} Larger $\zeta$ implies shorter run time, i.e., the the training time of MLP-5 plus the sampling time. The training time of SAE for feature extraction is not influenced by $\zeta$ and is excluded from this analysis.}
    	\label{fig:UTKFace_zeta_vs_total_time}
    \end{figure}
    
\end{remark}

\section{Experiments}\label{sec:experiment}
In this section, we will empirically evaluate the efficiency and the effectiveness of the proposed cDR-RS scheme in subsampling class-conditional GANs and CcGANs. We compare cDR-RS with six state-of-the-art sampling methods: Baseline (no subsampling or refining), GOLD \cite{mo2019mining}, Collab \cite{liu2020collaborative}, DRS \cite{azadi2018discriminator}, DDLS \cite{che2020your}, and DRE-F-SP+RS \cite{ding2020subsampling}. For sampling methods proposed for unconditional GANs, we implement them for each distinct class or regression label. For a fair comparison, extensive experiments are conducted on multiple benchmark datasets and network architectures, and diverse evaluation metrics are utilized to demonstrate the robust superiority of our cDR-RS method.

\subsection{Sampling from class-conditional GANs}\label{sec:experiment_classification}

{\setlength{\parindent}{0cm} \textbf{Experimental setup:}}
For class-conditional GANs, we conduct experiments on three image datasets: (i) CIFAR-10 \cite{krizhevsky2009learning}, (ii) CIFAR-100 \cite{krizhevsky2009learning}, and (iii) ImageNet-100 \cite{cao2017hashnet}, respectively. (i) CIFAR-10 consists of 60,000 ($32\times32$) RGB images uniformly from 10 classes. (ii) CIFAR-100 also includes 60,000 ($32\times32$) RGB images, but uniformly from 100 classes.  On both datasets, the overall number of training samples is 50,000, i.e., 5000 images per class on CIFAR-10 or 500 images per class for CIFAR-100. The remaining 10,000 samples are for test on CIFAR-10 or CIFAR-100. (iii) ImageNet-100 \cite{cao2017hashnet}, as a subset of ImageNet \cite{imagenet_cvpr09}, has 128,503 RGB images at $128\times 128$ resolution from 100 classes. In our experiment, we randomly split ImageNet-100 into a training set and a test set, where 10,000 images are for testing (on average 100 images per class) and the rest images are for training. 

\rev{In \Cref{tab:dataset_arch_samp_classification}, we show the investigated combinations of network architectures and sampling methods for the three datasets in this experiment.} 
On CIFAR-10, we train three class-conditional GANs: ACGAN \cite{odena2017conditional}, SNGAN \cite{miyato2018spectral}, and BigGAN \cite{brock2018large}. On CIFAR-100 and ImageNet-100, we only implement BigGAN because the training of ACGAN and SNGAN is unstable. When sampling from ACGAN on CIFAR-10, we test four candidate methods: Baseline, GOLD \cite{mo2019mining}, DRE-F-SP+RS \cite{ding2020subsampling}, and cDR-RS. When sampling from SNGAN on CIFAR-10 or BigGAN on all three datasets, we test six candidates: Baseline, Collab \cite{liu2020collaborative}, DRS \cite{azadi2018discriminator}, DDLS \cite{che2020your}, DRE-F-SP+RS \cite{ding2020subsampling}, and cDR-RS. Please note that Collab, DRS, and DDLS are inapplicable to ACGAN, and Baseline refers to sampling without subsampling or refining. Please refer to Supp.\ \ref{supp:details_of_cifar10}, \ref{supp:details_of_cifar100}, and \ref{supp:details_of_imagenet100} for details.

\begin{table}[htbp]
  \centering
  \caption{\rev{\textbf{The investigated combinations of datasets, network architectures, and sampling methods in the class-conditional GANs experiment.} We evaluate seven sampling methods in this experiment: Baseline (no subsampling or refining), GOLD \cite{mo2019mining}, Collab \cite{liu2020collaborative}, DRS \cite{azadi2018discriminator}, DDLS \cite{che2020your}, DRE-F-SP+RS \cite{ding2020subsampling}, and the proposed cDR-RS. However, when sampling from ACGAN \cite{odena2017conditional}, three candidate methods, i.e., Collab, DRS, and DDLS, are not used because of their inapplicability. When sampling from SNGAN \cite{miyato2018spectral} and BigGAN \cite{brock2018large}, GOLD is excluded because it is designed for ACGAN only.}}
    \begin{adjustbox}{width=0.75\textwidth}
    \begin{tabular}{lll}
    \hline\hline
    \textbf{Datasets} & \textbf{Architectures} & \textbf{Sampling methods} \\
    \hline
    \multicolumn{1}{l}{\multirow{3}[0]{*}{CIFAR-10}} & ACGAN & all methods except Collab, DRS, and DDLS \\
    \multicolumn{1}{l}{} & SNGAN & all methods except GOLD \\
    \multicolumn{1}{l}{} & BigGAN & all methods except GOLD \\
    \hline
    CIFAR-100 & BigGAN & all methods except GOLD \\
    \hline
    ImageNet-100 & BigGAN & all methods except GOLD \\
    \bottomrule
    \end{tabular}%
    \end{adjustbox}
  \label{tab:dataset_arch_samp_classification}%
\end{table}%

{\setlength{\parindent}{0cm} \textbf{Evaluation metrics:}} We evaluate the quality of fake images by \textit{Fr\'echet Inception Distance} (FID) \cite{heusel2017gans}, Intra-FID \cite{miyato2018cgans}, and \textit{Inception Score} (IS) \cite{salimans2016improved}. Intra-FID is an overall image quality metric, which computes FID separately for each class and reports the average FID score. A lower Intra-FID or FID score indicates better image quality or vice versa. Conversely, a larger Inception Score implies better image quality.

\begin{figure}[t] 
	\centering
	\includegraphics[width=0.7\linewidth]{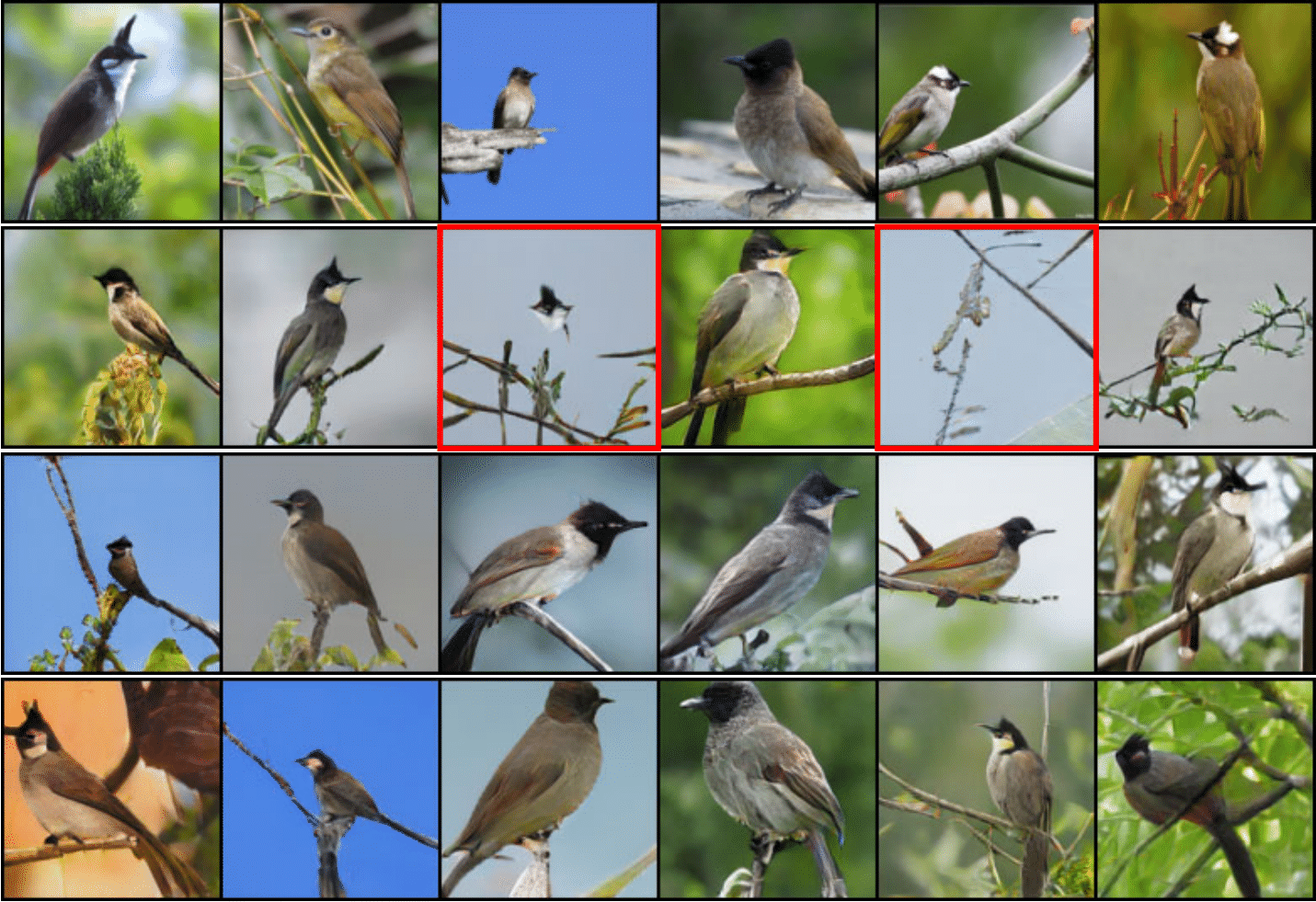} 
	\caption{\textbf{Some example images for the ``indigo bunting" class at $128\times 128$ resolution in the ImageNet-100 experiment.} The first row includes six real images. From the second row to the bottom, we show some example fake images generated from Baseline, DRE-F-SP+RS \cite{ding2020subsampling}, and the proposed cDR-RS, respectively. We often observe some fake images sampled by Baseline without recognizable birds (e.g., two images with red rectangles), while both DRE-F-SP+RS and cDR-RS can effectively remove such images. Please note that, as shown in \cref{fig:cover_fig} and \Cref{tab:ImageNet-100_efficiency_analysis}, although DRE-F-SP+RS and cDR-RS have similar effectiveness (similar Intra-FID, FID, and IS scores), cDR-RS requires much less storage or implementation time.}
	\label{fig:ImageNet-100_example_images}
\end{figure}

{\setlength{\parindent}{0cm}\textbf{Experimental results:}} We quantitatively compare the image quality of fake images sampled from class-conditional GANs with different candidate methods. For the CIFAR-10 experiment, we draw 10,000 fake images for each class by each sampling method. For the CIFAR-100 and ImageNet-100 experiments, we sample 1000 fake images for each class. Quantitative results are summarized in \Cref{tab:effectiveness_analysis_classification}. We can see, in all settings, the proposed cDR-RS and DRE-F-SP+RS perform comparably, and they substantially outperform other candidate methods by a large margin in terms of all three metrics. We also show in \cref{fig:ImageNet-100_example_images} some example fake images for the ``indigo bunting" class sampled by Baseline, DRE-F-SP+RS, and cDR-RS with real images for reference. Both DRE-F-SP+RS and cDR-RS can effectively remove some fake images with unrecognizable birds (marked by red rectangles). Although cDR-RS and DRE-F-SP+RS have similar effectiveness, their efficiency differs significantly. Our efficiency analysis on ImageNet-100, visualized in \cref{fig:cover_fig}(a) and summarized in  \Cref{tab:ImageNet-100_efficiency_analysis}, shows that cDR-RS only requires 1.19\% of the storage usage and 77\% of the implementation time consumed by DRE-F-SP+RS. Furthermore, since the cDRE-F-cSP is sample-based and does not rely on any properties of cGANs, cDR-RS is applicable to various cGAN architectures, making it more flexible than GOLD, Collab, DRS and DDLS.

\begin{table}[htbp]
	\centering
	\caption{\textbf{Effectiveness of different methods in sampling various class-conditional GANs in terms of Intra-FID, FID, and IS.} We evaluate the quality of 100,000 fake images (10,000 per class for CIFAR-10 or 1000 per class for CIFAR-100 and ImageNet-100) sampled from each method. The numbers in the parentheses are the standard deviations of the FIDs computed within each distinct class. The quality of real images from each datasets is also provided as references, where the Intra-FID and FID are computed between training and test samples while IS is computed in terms of test samples only. Please note that the test samples of ImageNet-100 are insufficient for computing a reliable Intra-FID. \textbf{In all settings, cDR-RS is either better than or comparable to DRE-F-SP+RS \cite{ding2020subsampling}, and these two methods substantially outperform the others. Compared with DRE-F-SP+RS, the advantage of cDR-RS in subsampling class-conditional GANs is its significantly improved efficiency (see Fig. \ref{fig:cover_fig} for efficiency comparisons). }}
	\begin{adjustbox}{width=0.5\textwidth}
		\begin{tabular}{rllll}
			\toprule
			& \textbf{Method} & \textbf{Intra-FID} $\downarrow$ & \textbf{FID} $\downarrow$ & \textbf{IS} $\uparrow$ \\
			\midrule
			\multicolumn{1}{c}{\multirow{20}[0]{*}{\begin{sideways}\textbf{CIFAR-10}\end{sideways}}} & Real Data & 0.681 (0.284) & 0.134 & 9.981 \\
			\cline{2-5}
			\multicolumn{1}{c}{} & \textbf{- ACGAN -} &       &       &  \\
			\multicolumn{1}{c}{} & Baseline & 3.688 (0.674) & 3.676 & 6.563 \\
			\multicolumn{1}{c}{} & GOLD \cite{mo2019mining} & 3.660 (0.682) & 3.682 & 6.603 \\
			\multicolumn{1}{c}{} & DRE-F-SP+RS \cite{ding2020subsampling} & 2.733 (0.519) & 2.518 & 8.122 \\
			\multicolumn{1}{c}{} & \textbf{cDR-RS (proposed)} & \textbf{2.656 (0.461)} & \textbf{2.209} & \textbf{8.294} \\
			\cline{2-5}
			\multicolumn{1}{c}{} & \textbf{- SNGAN -} &       &       &  \\
			\multicolumn{1}{c}{} & Baseline & 1.887 (0.265) & 1.426 & 8.947 \\
			\multicolumn{1}{c}{} & Collab \cite{liu2020collaborative} & 1.882 (0.274) & 1.433 & 8.941 \\
			\multicolumn{1}{c}{} & DRS \cite{azadi2018discriminator}  & 1.875 (0.281) & 1.417 & 8.958\\
			\multicolumn{1}{c}{} & DDLS \cite{che2020your}  & 1.769 (0.268) & 1.368 & 9.070 \\
			\multicolumn{1}{c}{} & DRE-F-SP+RS \cite{ding2020subsampling} & 1.164 (0.214) & 0.896 & 9.554 \\
			\multicolumn{1}{c}{} & \textbf{cDR-RS (proposed)} & \textbf{1.042 (0.233)} & \textbf{0.715} & \textbf{9.577} \\
			\cline{2-5}
			\multicolumn{1}{c}{} & \textbf{- BigGAN -} &       &       &  \\
			\multicolumn{1}{c}{} & Baseline & 0.998 (0.403) & 0.478 & 9.352 \\
			\multicolumn{1}{c}{} & Collab \cite{liu2020collaborative} & 0.993 (0.403) & 0.472 & 9.355 \\
			\multicolumn{1}{c}{} & DRS \cite{azadi2018discriminator}  & 1.034 (0.389) & 0.497 & 9.341 \\
			\multicolumn{1}{c}{} & DDLS \cite{che2020your} & 0.909 (0.380) & 0.469 & 9.436 \\
			\multicolumn{1}{c}{} & DRE-F-SP+RS \cite{ding2020subsampling} & 0.700 (0.251) & 0.385 & 9.605 \\
			\multicolumn{1}{c}{} & \textbf{cDR-RS (proposed)} & \textbf{0.594 (0.227)} & \textbf{0.282} & \textbf{9.648} \\
			
			\midrule
			
			\multicolumn{1}{c}{\multirow{9}[0]{*}{\begin{sideways}\textbf{CIFAR-100}\end{sideways}}} & Real Data & 16.976 (3.031) & 2.729 & 98.855 \\
			\cline{2-5}
			\multicolumn{1}{c}{} & \textbf{- BigGAN -} &       &       &  \\
			\multicolumn{1}{c}{} & Baseline & 19.752 (2.926) & 4.319 & 60.846 \\
			\multicolumn{1}{c}{} & Collab \cite{liu2020collaborative} & 19.799 (2.882) & 4.387 & 60.719 \\
			\multicolumn{1}{c}{} & DRS \cite{azadi2018discriminator}  & 19.742 (2.857) & 4.354 & 60.468 \\
			\multicolumn{1}{c}{} & DDLS \cite{che2020your} & 19.315 (2.678) & 4.011 & 64.095 \\
			\multicolumn{1}{c}{} & DRE-F-SP+RS \cite{ding2020subsampling} & \textbf{17.554 (1.559)} & \textbf{1.978} & \textbf{82.145} \\
			\multicolumn{1}{c}{} & \textbf{cDR-RS (proposed)} & 17.726 (1.618) & 1.993 & 80.393 \\
			
			\midrule
			
			\multicolumn{1}{c}{} & Real Data & --- & 2.979 & 89.771 \\
			\cline{2-5}
			\multicolumn{1}{c}{\multirow{7}[0]{*}{\begin{sideways}\textbf{ImageNet-100}\end{sideways}}} & \textbf{- BigGAN -} &       &       &  \\
			\multicolumn{1}{c}{} & Baseline & 35.023 (12.041) & 11.640 & 79.119 \\
			\multicolumn{1}{c}{} & Collab \cite{liu2020collaborative} & 35.178 (12.158) & 11.658 & 79.080 \\
			\multicolumn{1}{c}{} & DRS \cite{azadi2018discriminator}  & 33.997 (11.738) & 11.070 & 80.730 \\
			\multicolumn{1}{c}{} & DDLS \cite{che2020your} & 33.702 (11.604) & 10.833 & 80.918 \\
			\multicolumn{1}{c}{} & DRE-F-SP+RS \cite{ding2020subsampling} & \textbf{28.700 (9.526)} & 8.661 & \textbf{87.980} \\
			\multicolumn{1}{c}{} & \textbf{cDR-RS (proposed)} & 28.920 (9.070) & \textbf{8.328} & 87.824 \\
			\bottomrule
		\end{tabular}%
	\end{adjustbox}
	\label{tab:effectiveness_analysis_classification}%
\end{table}%

\begin{table}[htbp]
	\centering
	\caption{\textbf{Efficiency of different sampling methods on ImageNet-100 based on Two NVIDA V100.} For DRE-F-SP+RS and cDR-RS, the training time includes the time spent on the feature extractor training and the MLP-5 network training. Note that, with comparable effectiveness, cDR-RS only requires \textbf{1.19\%} of the storage usage and \textbf{77\%} of the implementation time spent by DRE-F-SP+RS. With much better performance, cDR-RS requires only \textbf{30.7\%} of the implementation time spent by DDLS. Collab and DRS require less storage space and implementation time, but they they are much less effective in sampling from cGANs (see \Cref{tab:effectiveness_analysis_classification}).}
	\begin{adjustbox}{width=0.8\textwidth}
		\begin{tabular}{lcccc}
			\toprule
			\textbf{Methods} & \textbf{ \begin{tabular}[c]{@{}c@{}} Total storage \\ usage (GB) \end{tabular} } & \textbf{ \begin{tabular}[c]{@{}c@{}} Total training \\ time (hours) \end{tabular} } & \textbf{ \begin{tabular}[c]{@{}c@{}} Total sampling \\ time (hours) \end{tabular} } & \textbf{ \begin{tabular}[c]{@{}c@{}} Total implementation \\ time (hours) \end{tabular} } \\
			\midrule
			Collab \cite{liu2020collaborative} & 0.13  & 5.58  & 5.23  & 10.81 \\
			DRS \cite{azadi2018discriminator}  & 0.13  & 1.47  & 0.75  & 2.22 \\
			DDLS \cite{che2020your} & 0.13     & 0     & 218.05 & 218.05 \\
			DRE-F-SP+RS \cite{ding2020subsampling} & 38.94 & 84.41 & 2.43  & 86.84 \\
			\textbf{Proposed: cRS-RS (Filter)} & 0.75  & 65.69  & 1.18  & 66.87 \\
			\bottomrule
		\end{tabular}%
	\end{adjustbox}
	\label{tab:ImageNet-100_efficiency_analysis}%
\end{table}%

\subsection{Sampling from CcGANs}\label{sec:experiment_regression}

{\setlength{\parindent}{0cm} \textbf{Experimental setup:}} Besides class-conditional GANs, we evaluate the performances of candidate methods in sampling from CcGANs. We experiment on the UTKFace \cite{utkface} and RC-49 datasets \cite{ding2021ccgan, ding2020continuous}, which are two benchmark datasets for CcGANs. UTKFace is an RGB human face image dataset with ages as regression labels. We use the pre-processed UTKFace dataset \cite{ding2021ccgan}, consisting of 14,760 RGB images at $64\times 64$ resolution with ages in [1, 60]. The number of images in UTKFace ranges from 50 to 1051 for different ages, and all images are used for training. RC-49 consists of 44,051 RGB images at $64\times 64$ resolution for 49 types of chairs. Each type of chair contains 899 images labeled by 899 distinct yaw rotation angles from $0.1^{\circ}$ to $89.9^{\circ}$ with a step size of $0.1^{\circ}$. We first choose yaw angles with odd numbers as the last digit for training. Then, for each chosen angle, we randomly select 25 images to construct a training set. The final training set of RC-49 includes 11250 images and 450 distinct angles.  All 44,051 RC-49 images are used in the experiment evaluation.

We follow the official implementation of CcGAN (SVDL+ILI) in \cite{ding2021ccgan, ding2020continuous}. GOLD is not taken as a baseline method because of its incompatibility in subsampling CcGANs. DRE-F-SP+RS is tested on UTKFace but excluded from the RC-49 experiment because it cannot subsample images that are generated conditional on unseen labels. DDLS is also not tested in the RC-49 experiment due to its extremely long sampling time. When implementing DRE-F-SP+RS, we use the SAE proposed in \Cref{sec:cDRE-F-cSP} to extract features. Please refer to Supp.\ \ref{supp:details_of_utkface} and \ref{supp:details_of_rc49} for details.

{\setlength{\parindent}{0cm} \textbf{Evaluation metrics:}}
Following \cite{ding2021ccgan, ding2020continuous}, we adopt four evaluation metrics: Intra-FID \cite{miyato2018cgans}, \textit{Naturalness Image Quality Evaluator} (NIQE) \cite{mittal2012making}, Diversity \cite{ding2021ccgan, ding2020continuous}, and Label Score \cite{ding2021ccgan, ding2020continuous}. Intra-FID is taken as the overall image quality metric, which computes FID separately at each evaluation angle and report the average score (along with the standard deviation). Diversity measures the diversity of fake images generated conditional on a given label in terms of some categorical properties of fake images (i.e., races for UTKFace and chair types for RC-49). Label Score evaluates label consistency, measuring the discrepancy between actual and conditioning labels of fake images. Quantitatively, we prefer smaller Intra-FID, NIQE and Label Score indices but larger Diversity values. Please see Supp.\ \ref{supp:details_of_utkface} and \ref{supp:details_of_rc49} for detailed definitions.

\begin{figure}[htbp]
	\centering
	\includegraphics[width=0.8\linewidth]{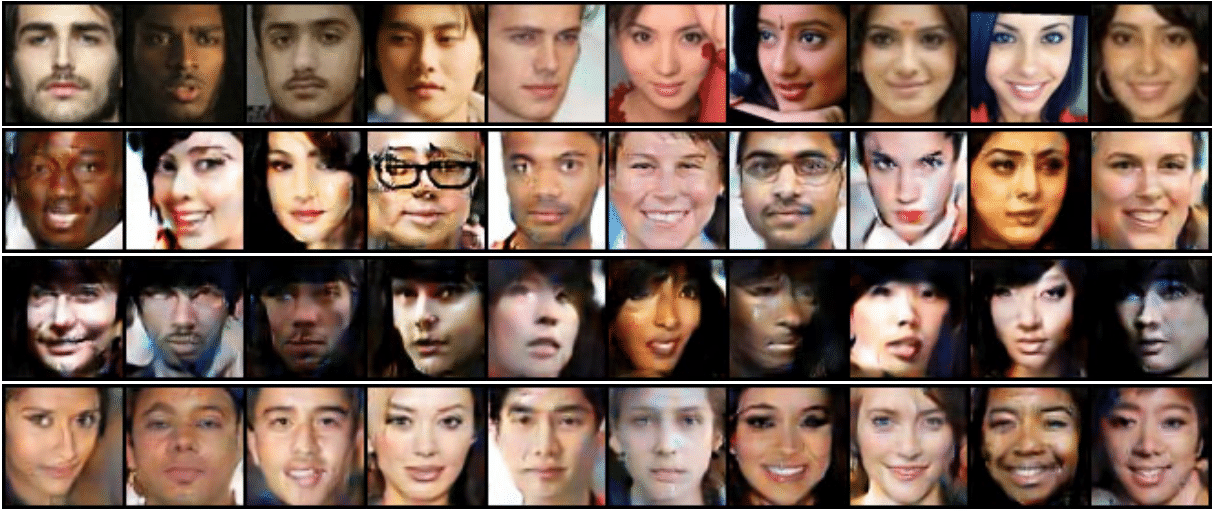}  
	\caption{\textbf{Some example images for age 24 at $64\times 64$ resolution in the UTKFace experiment.} The first row includes ten real images. From the second row to the bottom, we show some example fake images generated from Baseline, DRE-F-SP+RS \cite{ding2020subsampling}, and the proposed cDR-RS, respectively. By comparing Row 2 and 4, we can see cDR-RS can effectively improve the visual quality. By contrast, DRE-F-SP+RS worsens the visual quality.}
	\label{fig:UTKFace_example_images}
\end{figure}

\begin{figure}[htbp]
	\centering
	\includegraphics[width=0.6\linewidth]{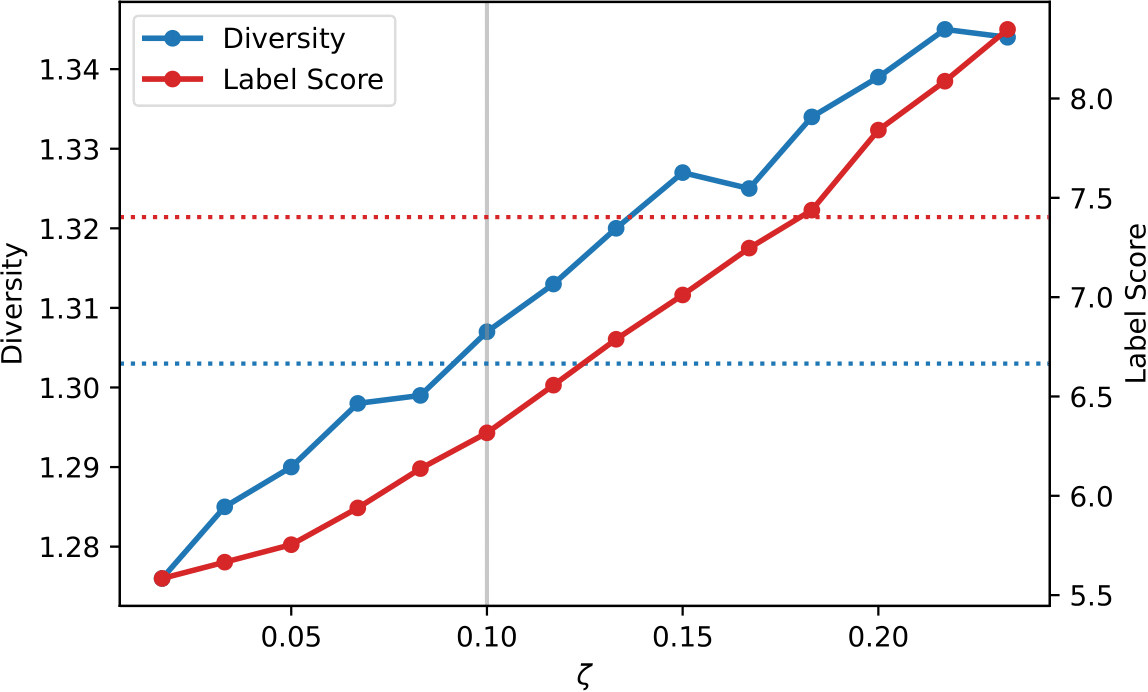}  
	\caption{\textbf{The effects of $\zeta$ in the filtering scheme of cDR-RS on the relation between Diversity and Label Score of CcGAN-generated samples in the UTKFace experiment.} The dotted black and red lines stand for Diversity and Label Score of Baseline in \Cref{tab:effectiveness_analysis_regression}, respectively. The vertical grey line specifies the $\zeta$ used in \Cref{tab:effectiveness_analysis_regression}. We can see both Label Score and Diversity increase as $\zeta$ increases. If we prefer higher label consistency, we can decrease $\zeta$. Oppositely, if we prefer higher image diversity, we can increase $\zeta$. No matter what the preference is, when choosing $\zeta$, we should ensure that the corresponding Label Score is below the red dotted line while Diversity is above the black dotted line. In that case, both label consistency and image diversity can be improved. A rule of thumb for the parameter selection is provided in \Cref{rmk:rule_of_thumb_for_zeta}.}
	\label{fig:UTKFace_effect_of_kappa}
\end{figure}

{\setlength{\parindent}{0cm}\textbf{Experimental results:}} {\color{black}
We quantitatively compare the performance of different candidate methods in sampling from CcGANs. For the UTKFace experiment, we sample 1000 fake images for each age by each method. For the RC-49 experiment, we sample 200 fake images for each of 899 distinct angles by each method (449 angles are unseen in the training set). 

We compare the effectiveness of candidate methods based on multiple evaluation metrics in \Cref{tab:effectiveness_analysis_regression}. We can see Collab has little effect when sampling from CcGANs. DRS can improve NIQE in both experiments, but it fails to simultaneously improve Diversity and Label Score. DDLS improves visual quality in the UTKFace experiment, but it sacrifices Diversity for a slightly lower Label Score. The sampling time (11.24 hours) spent by DDLS is also much longer than that for the other methods, making DDLS infeasible for the RC-49 experiment. DRE-F-SP+RS performs worst among all methods, and it does not apply to RC-49. The proposed cDR-RS with the filtering scheme, denoted by cDR-RS (Filter), outperforms all other candidate methods on both datasets regarding Intra-FID and Label Score. We also show in \cref{fig:UTKFace_example_images} some example fake images for age 24 sampled by Baseline, DRE-F-SP+RS, and cDR-RS (Filter) with real images as reference. \cref{fig:UTKFace_example_images} demonstrates the effectiveness of cDR-RS (Filter) and the failure of DRE-F-SP+RS. In \Cref{tab:effectiveness_analysis_regression}, we also show the performance of cDR-RS without the filtering scheme, i.e., cDR-RS (No Filter). We can see, cDR-RS (No filter) cannot effectively solve the label inconsistency problem and even makes it worse. Thus, this observation validates the effectiveness of the proposed filtering scheme. In addition, we conduct an ablation study on UTKFace to analyze the effects of $\zeta$ of the filtering scheme, and this analysis is visualized in \cref{fig:UTKFace_effect_of_kappa}. 

The efficiency analysis of candidate methods on UTKFace is shown in \Cref{tab:UTKFace_efficiency_analysis}. Collab and DRS request low storage space and very little implementation time. Oppositely, DDLS spends a very long implementation time, and DRE-F-SP+RS requests a lot of storage space. cDR-RS requires more storage space and implementation time than Collab and DRS do, but it is much more efficient than DDLS and DRE-F-SP+RS.

In terms of the effectiveness and efficiency analysis, we can conclude that, although Collab and DRS are very efficient, they fail to improve CcGANs effectively. DRE-F-SP+RS is inefficient and ineffective in sampling from CcGANs. DDLS is not a practical sampling method for CcGANs because its implementation takes too long. Unlike other candidate methods, cDR-RS requires only reasonable computational resources but leads to substantial performance improvements.


}

\begin{figure}[!ht]
	\centering
	\includegraphics[width=0.6\linewidth]{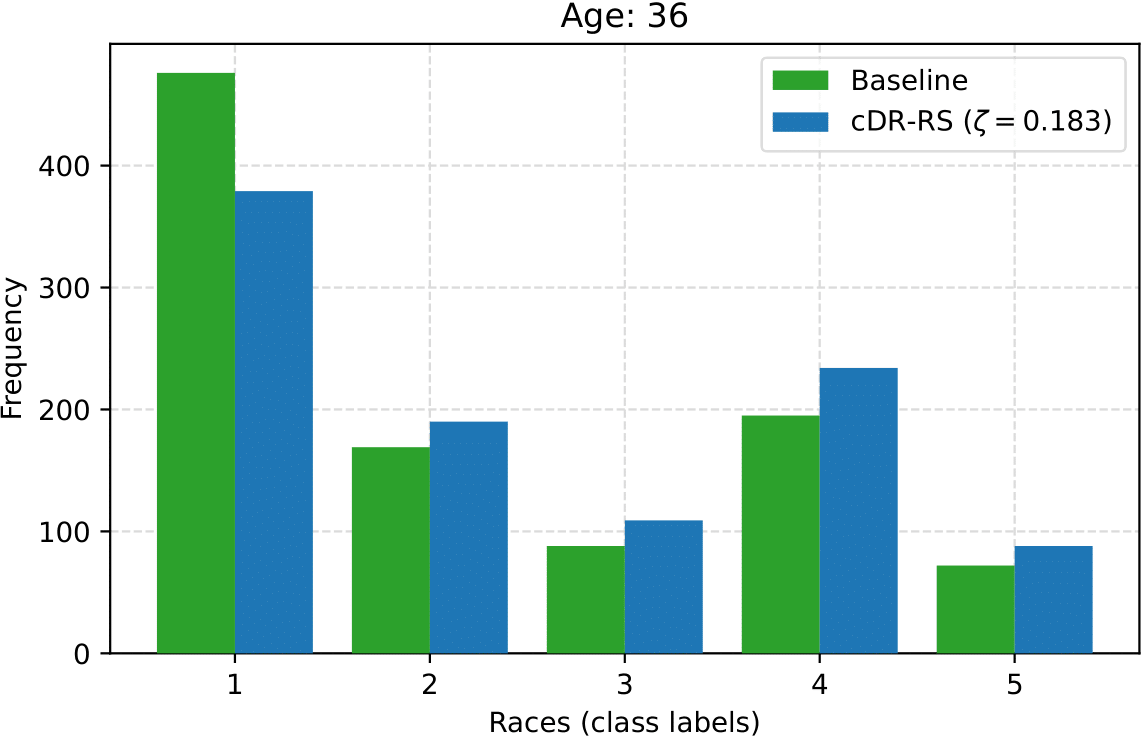}  
	\caption{\textbf{The distributions of fake images sampled from Baseline and cDR-RS, respectively, at age 36 over 5 races in the UTKFace experiment.} After subsampling by cDR-RS, the distribution of fake images for 5 races is more balanced, resulting in higher diversity.}
	\label{fig:UTKFace_diversity_increase_analysis}
\end{figure}

{\setlength{\parindent}{0cm}\textbf{Why the Diversity score is improved?}} 
It may be counter-intuitive that the Diversity score is increased when cDR-RS may reject some generated images. \cref{fig:UTKFace_diversity_increase_analysis} provides an explanation by visualizing the distributions of 1000 fake images sampled from Baseline and cDR-RS, respectively, for age 36 over 5 races in the UTKFace experiment. To make the illustration clearer, we increase $\zeta$ to $0.183$, so that the Label Scores of Baseline and cDR-RS are comparable, and the improvement caused by cDR-RS focuses on Diversity. Please note that, in the evaluation of the UTKFace experiment, Diversity is defined as the average entropy of the races of fake images that are predicted by a pre-trained classification CNN (races are taken as class labels). Therefore, a more balanced race distribution implies higher Diversity. From \cref{fig:UTKFace_diversity_increase_analysis}, we can see that, after applying cDR-RS, the frequency of Race 1 decreases while the frequencies for other races increase. Therefore, the Diversity score is improved.


\begin{table}[htbp]
	\centering
	\caption{\textbf{Effectiveness analysis of different sampling methods with CcGANs (SVDL+ILI) and two regression datasets in terms of Intra-FID, NIQE, Diversity, and Label Score.} Values in the parentheses represent the standard deviation of evaluation scores reported at each distinct regression label. The proposed cDR-RS substantially outperforms other candidate methods on both datasets.}
	\begin{adjustbox}{width=0.75\textwidth}
		\begin{tabular}{lcccc}
			\toprule
			\textbf{Method} & \textbf{Intra-FID} $\downarrow$ & \textbf{NIQE} $\downarrow$ & \textbf{Diversity} $\uparrow$ & \textbf{Label Score} $\downarrow$ \\
			\midrule
			\textbf{- UTKFace -} &  &  &  &  \\
			Baseline & 0.457 (0.171) & 1.722 (0.172) & 1.303 (0.170) & 7.403 (5.956) \\
			Collab \cite{liu2020collaborative} & 0.457 (0.168) & 1.722 (0.170) & 1.298 (0.180) & 7.420 (6.030) \\
			DRS \cite{azadi2018discriminator} & 0.455 (0.169) & 1.707 (0.176) & 1.282 (0.185) & 7.487 (6.105) \\
			DDLS \cite{che2020your} & 0.445 (0.166) & 1.712 (0.172)  & 1.288 (0.180) & 7.202 (5.888) \\
			DRE-F-SP+RS \cite{ding2020subsampling} & 0.588 (0.657) & 1.724 (0.173) & 1.293 (0.184) & 7.462 (6.046) \\
			\hdashline
			Proposed: cDR-RS (No filter) & 0.443 (0.183) & \textbf{1.703 (0.168)} & \textbf{1.348 (0.139)} & 7.528 (6.125) \\
			\textbf{Proposed: cDR-RS (Filter)} & \textbf{0.430 (0.176)} & 1.708 (0.169) & 1.307 (0.208) & \textbf{6.317 (5.026)} \\
			\midrule
			\textbf{- RC-49 -} &  &  &  &  \\
			Baseline & 0.389 (0.096) & 1.759 (0.168) & 2.950 (0.067) & 1.938 (1.484) \\
			Collab \cite{liu2020collaborative} & 0.389 (0.095) & 1.762 (0.174) & 2.952 (0.067) & 1.939 (1.487) \\
			DRS \cite{azadi2018discriminator} & 0.360 (0.097) & \textbf{1.751 (0.149)} & 2.853 (0.132) & 1.855 (1.404) \\
			\hdashline
			\textbf{Proposed: cDR-RS (Filter)} & \textbf{0.334 (0.095)} & 1.756 (0.163) & \textbf{3.048 (0.090)} & \textbf{1.114 (0.885)} \\
			\bottomrule
		\end{tabular}%
	\end{adjustbox}
	\label{tab:effectiveness_analysis_regression}%
\end{table}%

\begin{table}[htbp]
	\centering
	\caption{\textbf{Efficiency analysis of different sampling methods on UTKFace based on One NVIDA V100.} For DRE-F-SP+RS and cDR-RS, the training time includes the time spent on the SAE training and the MLP-5 network training. cDR-RS is more efficient than DRE-F-SP+RS and DDLS in terms of either storage usage or implementation time. Collab and DRS requires less storage space and implementation time, but they do not have significant enough effectiveness in sampling from CcGANs (see \Cref{tab:effectiveness_analysis_regression}).}
	\begin{adjustbox}{width=0.8\textwidth}
		\begin{tabular}{lccccc}
			\toprule
			\textbf{Methods} & \textbf{ \begin{tabular}[c]{@{}c@{}} Total storage \\ usage (MB) \end{tabular} } & \textbf{ \begin{tabular}[c]{@{}c@{}} Total training \\ time (hours) \end{tabular} } & \textbf{ \begin{tabular}[c]{@{}c@{}} Total sampling \\ time (hours) \end{tabular} } & \textbf{ \begin{tabular}[c]{@{}c@{}} Total implementation \\ time (hours) \end{tabular} } \\
			\midrule
			Collab \cite{liu2020collaborative} & 82.8  & 1.05  & 0.27  & 1.32 \\
			DRS \cite{azadi2018discriminator}  & 82.8  & 0.16  & 0.13  & 0.29 \\
			DDLS \cite{che2020your} & 82.8  & 0  & 11.24 & 11.24 \\
			DRE-F-SP+RS \cite{ding2020subsampling} & 6,671 & 1.92 & 5.69  & 7.61 \\
			\textbf{Proposed: cRS-RS (Filter)} & 303  & 1.99  & 0.49  & 2.48 \\
			\bottomrule
		\end{tabular}%
	\end{adjustbox}
	\label{tab:UTKFace_efficiency_analysis}%
\end{table}%

\section{Conclusion} \label{sec:conclusion}
In this work, we have presented a novel conditional subsampling scheme to improve the image quality of fake images from conditional GANs, with conditioning on a class or a continuous variable. First, we propose novel conditional extensions of density ratio estimation (cDRE) in the feature space and the Softplus loss function (cSP). Then, we learn the conditional density ratio model through an MLP network. Also, we derive the error bound of a conditional density ratio model trained with the proposed cSP loss. A novel filtering scheme is also proposed in subsampling CcGANs to improve the label consistency. Finally, we validate the superiority of the proposed subsampling framework with extensive experiments in sampling multiple conditional GAN models on four benchmark datasets with diverse evaluation metrics.

\bibliographystyle{elsarticle-num} 
\bibliography{cdre_ref}

\clearpage
\newpage
\appendix

\section*{Supplementary Material}
\addcontentsline{toc}{section}{Supplementary Material}
\renewcommand{\thesection}{S.\arabic{section}} 
\renewcommand{\thesubsection}{\thesection.\arabic{subsection}}
\renewcommand\thefigure{\thesection.\arabic{figure}}
\renewcommand\thetable{\thesection.\arabic{table}}
\renewcommand{\theequation}{S.\arabic{equation}}
\renewcommand{\thetheorem}{S.\arabic{theorem}} 
\renewcommand{\thedefinition}{S.\arabic{definition}} 
\renewcommand{\thelemma}{S.\arabic{lemma}} 
\renewcommand{\theremark}{S.\arabic{remark}}

\section{The Proof of Theorems \ref{thm:error_bound}}\label{supp:proofs}
\begin{proof}
	Following \cite{ding2020subsampling}, we decompose $\mathcal{L}_c(\hat{\psi})-\mathcal{L}_c(\psi^*)$ as follows:
	
	\begin{align}
		&\mathcal{L}_c(\hat{\psi})-\mathcal{L}_c(\psi^*) \nonumber\\
		=& \mathcal{L}_c(\hat{\psi}) - \widehat{\mathcal{L}}_c(\hat{\psi}) + \widehat{\mathcal{L}}_c(\hat{\psi}) - \widehat{\mathcal{L}}_c(\tilde{\psi}) + \widehat{\mathcal{L}}_c(\tilde{\psi}) - \mathcal{L}_c(\tilde{\psi})  + \mathcal{L}_c(\tilde{\psi}) - \mathcal{L}_c(\psi^*) \nonumber\\
		& (\text{Since } \widehat{\mathcal{L}}_c(\hat{\psi}) - \widehat{\mathcal{L}}_c(\tilde{\psi})\leq 0) \nonumber\\
		\leq& \mathcal{L}_c(\hat{\psi}) - \widehat{\mathcal{L}}_c(\hat{\psi}) + \widehat{\mathcal{L}}_c(\tilde{\psi}) - \mathcal{L}_c(\tilde{\psi}) + \mathcal{L}_c(\tilde{\psi}) - \mathcal{L}_c(\psi^*)\nonumber\\
		\leq& 2\sup_{\psi\in\Psi}\left| \widehat{\mathcal{L}}_c(\psi) - \mathcal{L}_c(\psi) \right| + \mathcal{L}_c(\tilde{\psi}) - \mathcal{L}_c(\psi^*).
		\label{eq:first_decompose}
	\end{align}
	The second term in \cref{eq:first_decompose} is a constant which implies an inevitable error. The first term can be bounded as follows:
	
	\begin{align}
		&\sup_{\psi\in\Psi}\left| \widehat{\mathcal{L}}_c(\psi) - \mathcal{L}_c(\psi) \right| \nonumber\\
		\leq & \sup_{\psi\in\Psi} \left| \mathbb{E}_{(\bm{h},y)\sim q_g(\bm{h},y)}\left[ \sigma(\psi(\bm{h}|y))\psi(\bm{h}|y) - \eta(\psi(\bm{h}|y)) \right]  - \frac{1}{N^g}\sum_{i=1}^{N^g}\left[ \sigma(\psi(\bm{h}^g_i|y^g_i))\psi(\bm{h}^g_i|y^g_i) - \eta(\psi(\bm{h}^g_i|y^g_i)) \right]  \right|\nonumber\\
		& + \left| \mathbb{E}_{(\bm{h},y)\sim q_r(\bm{h},y)}\left[ \sigma(\psi(\bm{h}|y))\right] - \frac{1}{N^r}\sum_{i=1}^{N^r}\sigma(\psi(\bm{h}^r_i|y^r_i))  \right|.
		\label{eq:second_decompose}
	\end{align}
	
	Because of assumptions (i)-(iv), based on the uniform law of large number \cite{noteULLN}, for $\forall \epsilon>0$,
	
	\begin{equation*}
		\scalemath{1}{
			\begin{aligned}
				\lim_{N^g\rightarrow\infty} & P\left\{  \sup_{\psi\in\Psi} \left|\vphantom{\frac{1}{N^g}} \mathbb{E}_{(\bm{h},y)\sim q_g(\bm{h},y)}\left[ \sigma(\psi(\bm{h}|y))\psi(\bm{h}|y) - \eta(\psi(\bm{h}|y)) \right] \right.\right. \\
				& - \left.\left. \frac{1}{N^g}\sum_{i=1}^{N^g}\left[ \sigma(\psi(\bm{h}^g_i|y^g_i))\psi(\bm{h}^g_i|y^g_i) - \eta(\psi(\bm{h}^g_i|y^g_i)) \right]  \vphantom{\frac{1}{N^g}}\right|\right\} = 0.
			\end{aligned}
		}
	\end{equation*}
	
	Based on this limit, we can derive an upper bound of the first term of \cref{eq:second_decompose} as follows. Since we can generate infinite fake images from a trained cGAN, $N^g$ is large enough. Let $\epsilon={1}/{2N^g}$, $\forall \delta_1\in(0,1)$ with probability at least $1-\delta_1$, 
	
	\begin{equation}
		\scalemath{1}{
			\begin{aligned}
				& \sup_{\psi\in\Psi} \left|\vphantom{\frac{1}{N^g}} \mathbb{E}_{(\bm{h},y)\sim q_g(\bm{h},y)}\left[ \sigma(\psi(\bm{h}|y))\psi(\bm{h}|y) - \eta(\psi(\bm{h}|y)) \right] \right. \\
				& - \left. \frac{1}{N^g}\sum_{i=1}^{N^g}\left[ \sigma(\psi(\bm{h}^g_i|y^g_i))\psi(\bm{h}^g_i|y^g_i) - \eta(\psi(\bm{h}^g_i|y^g_i)) \right]  \vphantom{\frac{1}{N^g}}\right| \leq \frac{1}{2N^g}.
				\label{eq:bound_first_term}
			\end{aligned}
		}
	\end{equation}
	
	The second term of \cref{eq:second_decompose} can be bounded based on Lemma 1 and Theorem 2 (the Rademacher bound \cite{lafferty2010}) in \cite{ding2020subsampling} as follows: $\forall \delta_2\in(0,1)$ with probability at least $1-\delta_2$,
	
	\begin{align}
		&\left| \mathbb{E}_{(\bm{h},y)\sim q_r(\bm{h},y)}\left[ \sigma(\psi(\bm{h}|y))\right] - \frac{1}{N^r}\sum_{i=1}^{N^r}\sigma(\psi(\bm{h}^r_i|y^r_i))  \right|\nonumber\\
		\leq&2\hat{\mathcal{R}}_{q_r(\bm{h},y),N^r}(\sigma\circ\Psi) + \sqrt{\frac{4}{N^r}\log\left(\frac{2}{\delta_2}\right)}\nonumber\\
		\leq&\frac{1}{2}\hat{\mathcal{R}}_{q_r(\bm{h},y),N^r}(\Psi) + \sqrt{\frac{4}{N^r}\log\left(\frac{2}{\delta_2}\right)}.
		\label{eq:bound_second_term}
	\end{align}
	Let $\delta=\max\{\delta_1, \delta_2\}$ and $\delta^\prime=\delta$, based on \cref{eq:bound_first_term} and \eqref{eq:bound_second_term}, we can derive Eq.\ \eqref{eq:error_bound}.
\end{proof}

\section{Resources for Implementing cGANs and Sampling Methods}\label{supp:resources}

To implement ACGAN, we refer to \url{https://github.com/sangwoomo/GOLD}.

To implement SNGAN, we refer to \url{https://github.com/christiancosgrove/pytorch-spectral-normalization-gan} and \url{https://github.com/pfnet-research/sngan_projection}. 

To implement BigGAN, we refer to \url{https://github.com/ajbrock/BigGAN-PyTorch}. 

To implement CcGANs, we refer to \url{https://github.com/UBCDingXin/improved_CcGAN}.

To implement GOLD, we refer to \url{https://github.com/sangwoomo/GOLD}.

To implement Collab, we refer to \url{https://github.com/YuejiangLIU/pytorch-collaborative-gan-sampling}.

To implement DRS and DRE-F-SP+RS, we refer to \url{https://github.com/UBCDingXin/DDRE_Sampling_GANs}. 

To implement DDLS, we refer to \url{https://github.com/JHpark1677/CGAN-DDLS} and \url{https://github.com/Daniil-Selikhanovych/ebm-wgan/blob/master/notebook/EBM_GAN.ipynb}. 

\section{More Details of Experiments on CIFAR-10}\label{supp:details_of_cifar10}

In the CIFAR-10 experiment, we first train ACGAN, SNGAN, and BigGAN with setups shown as follows. We train ACGAN for 100,000 iterations with batch size 512. We train SNGAN for 50,000 iterations with batch size 512. BigGAN is trained for 39,000 with batch size 512. We use the architecture described in Figure 15 of \cite{brock2018large} for BigGAN.

To implement Collab, following \cite{liu2020collaborative}, we conduct discriminator shaping for 5000 and 2500 iterations for SNGAN and BigGAN, respectively. We do the refinement 30 times in a middle layer of SNGAN and 20 times in a middle layer of BigGAN. The step size of refinement is set as 0.1 for both SNGAN and BigGAN. 

To implement DDLS, we run the Langevin dynamics procedure for SNGAN and BigGAN within each class with step size $10^{-4}$ up to 1000 iterations.

To implement DRE-F-SP+RS, we first train the specially designed ResNet-34 on the training set for 350 epochs with the SGD optimizer, initial learning rate 0.1 (decayed at epoch 150 and 250 with factor 0.1), weight decay $10^{-4}$, and batch size 256. Ten MLP-5 models for modeling the density ratio function within each class are trained on the training set with the Adam optimizer \cite{kingma2014adam}, initial learning rate $10^{-4}$ (decayed at epoch 100 and 250), batch size 256, 400 epochs, and $\lambda=10^{-2}$. The network architecture of MLP-5 is shown in \Cref{tab:cifar10_MLP5}. 

To implement cDR-RS, we use the specially designed ResNet-34 in the implementation of DRE-F-SP+RS to extract features from images. The MLP-5 to model the conditional density ratio function is shown in \Cref{tab:cifar10_cMLP5}. It is trained with the Adam optimizer \cite{kingma2014adam}, initial learning rate $10^{-4}$ (decayed at epoch 80 and 150), batch size 256, 200 epochs, and $\lambda=10^{-2}$.

For a more accurate evaluation, we do not use Inception-V3 \cite{szegedy2016rethinking} that was pre-trained on ImageNet \cite{imagenet_cvpr09} to compute Intra-FID, FID, and IS. Instead, following \cite{ding2020subsampling}, we train Inception-V3 from scratch on CIFAR-10 to evaluate fake images. 

Please refer to our codes for more detailed setups such as network architectures and hyperparameter settings.

\begin{table}[h]
	\centering
	\caption{The 5-layer MLP for DRE in feature space for CIFAR-10 and CIFAR-100.}
		\begin{tabular}{c}
			\toprule
			Extracted feature $\bm{h}\in \mathbbm{R}^{3072}$ \\
			\hline
			fc$\rightarrow 2048$, GN (8 groups), ReLU, Dropout($p=0.5$) \\\hline
			fc$\rightarrow 1024$, GN (8 groups), ReLU, Dropout($p=0.5$) \\\hline
			fc$\rightarrow 512$, GN (8 groups), ReLU, Dropout($p=0.5$) \\\hline
			fc$\rightarrow 256$, GN (8 groups), ReLU, Dropout($p=0.5$) \\\hline
			fc$\rightarrow 128$, GN (8 groups), ReLU, Dropout($p=0.5$) \\\hline
			fc$\rightarrow 1$, ReLU \\
			\bottomrule
		\end{tabular}%
	\label{tab:cifar10_MLP5}%
\end{table}%

\begin{table}[h]
	\centering
	\caption{The 5-layer MLP for cDRE in feature space for CIFAR-10 and CIFAR-100. The embedded class label is appended to the extracted feature $\bm{h}$.}
		\begin{tabular}{c}
			\toprule
			Input: extracted feature $\bm{h}\in \mathbbm{R}^{3072}$ \\
			and embedded class label $y\in \mathbbm{R}^{C}$, \\
			where $C=10$ for CIFAR-10 and $C=100$ for CIFAR-100 \\
			\hline
			Concatenate $[\bm{h},y]\in \mathbbm{R}^{3082}$\\
			\hline
			fc$\rightarrow 2048$, GN (8 groups), ReLU, Dropout($p=0.5$) \\\hline
			fc$\rightarrow 1024$, GN (8 groups), ReLU, Dropout($p=0.5$) \\\hline
			fc$\rightarrow 512$, GN (8 groups), ReLU, Dropout($p=0.5$) \\\hline
			fc$\rightarrow 256$, GN (8 groups), ReLU, Dropout($p=0.5$) \\\hline
			fc$\rightarrow 128$, GN (8 groups), ReLU, Dropout($p=0.5$) \\\hline
			fc$\rightarrow 1$, ReLU \\
			\bottomrule
		\end{tabular}%
	\label{tab:cifar10_cMLP5}%
\end{table}%

\section{More Details of Experiments on CIFAR-100}\label{supp:details_of_cifar100}

In the CIFAR-100 experiment, we only test candidate sampling methods on BigGAN because both ACGAN and SNGAN are unstable on CIFAR-100 and somehow suffer from the model collapse problem \cite{gulrajani2017improved}. BigGAN is trained for 38,000 iterations with batch size 512 on the training set of CIFAR-100. We use the architecture described in Figure 15 of \cite{brock2018large} for BigGAN. We also adopt DiffAugment \cite{zhao2020differentiable} (a data augmentation method for the GAN training with limited data) to improve the performance of BigGAN. The strongest data augmentation policy, ``color,translation,cutout", is used for DiffAugment.

To implement Collab, following \cite{liu2020collaborative}, we conduct discriminator shaping for 2500 iterations for BigGAN. We do the refinement 30 times in a middle layer of the generator network of BigGAN. The step size of refinement is set as 0.1. 

To implement DDLS, we run the Langevin dynamics procedure for BigGAN within each class with step size $10^{-4}$ up to 1000 iterations.

To implement DRE-F-SP+RS, we first train the specially designed ResNet-34 on the training set for 350 epochs with the SGD optimizer, initial learning rate 0.1 (decayed at epoch 150 and 250 with factor 0.1), weight decay $10^{-4}$, and batch size 256. Ten MLP-5 models for modeling the density ratio function within each class are trained on the training set with the Adam optimizer \cite{kingma2014adam}, initial learning rate $10^{-4}$ (decayed at epoch 100 and 250), batch size 256, 400 epochs, and $\lambda=10^{-2}$. The network architecture of MLP-5 is shown in \Cref{tab:cifar10_MLP5}. 

To implement cDR-RS, we use the specially designed ResNet-34 in the implementation of DRE-F-SP+RS to extract features from images. The MLP-5 to model the conditional density ratio function is shown in \Cref{tab:cifar10_cMLP5}. It is trained with the Adam optimizer \cite{kingma2014adam}, initial learning rate $10^{-4}$ (decayed at epoch 80 and 150), batch size 256, 200 epochs, and $\lambda=10^{-2}$.

For a more accurate evaluation, we do not use Inception-V3 \cite{szegedy2016rethinking} that was pre-trained on ImageNet \cite{imagenet_cvpr09} to compute Intra-FID, FID, and IS. Instead, following \cite{ding2020subsampling}, we train Inception-V3 from scratch on CIFAR-100 to evaluate fake images. 

Please refer to our codes for more detailed setups such as network architectures and hyperparameter settings.

\section{More Details of Experiments on ImageNet-100}\label{supp:details_of_imagenet100}

\subsection{Setups of training, sampling, and evaluation}

In the ImageNet-100 dataset, we implement BigGAN with the BigGAN-deep architecture described in Figure 16 of \cite{brock2018large}. BigGAN is trained for 96,000 iterations with batch size 1024. We also adopt DiffAugment \cite{zhao2020differentiable} (a data augmentation method for the GAN training with limited data) to improve the performance of BigGAN. The strongest data augmentation policy, ``color,translation,cutout", is used for DiffAugment.

To implement Collab, following \cite{liu2020collaborative}, we conduct discriminator shaping for 3000 iterations for BigGAN. We do the refinement 16 times in a middle layer of the generator network of BigGAN. The step size of refinement is set as 0.5. 

To implement DRS, we fine-tune the discriminator of BigGAN for 5 epochs with batch size 128.  

To implement DDLS, we run the Langevin dynamics procedure for BigGAN within each class with step size $10^{-4}$ up to 1000 iterations.

To implement DRE-F-SP+RS, we first train the specially designed ResNet-34 on the training set for 350 epochs with the SGD optimizer, initial learning rate 0.1 (decayed at epoch 150 and 250 with factor 0.1), weight decay $10^{-4}$, and batch size 128. Ten MLP-5 models for modeling the density ratio function within each class are trained on the training set with the Adam optimizer \cite{kingma2014adam}, initial learning rate $10^{-4}$ (decayed at epoch 100 and 250), batch size 256, 400 epochs, and $\lambda=10^{-2}$. The network architecture of MLP-5 is similar to \Cref{tab:cifar10_MLP5}. Please note that, when implementing DRE-F-SP+RS, we use more epochs than cDR-RS does (400 epochs vs 200 epochs) to ensure that all density ratio models are well-trained. 

To implement cDR-RS, we use the specially designed ResNet-34 in the implementation of DRE-F-SP+RS to extract features from images. The MLP-5 to model the conditional density ratio function is similar to \Cref{tab:cifar10_cMLP5}. It is trained with the Adam optimizer \cite{kingma2014adam}, initial learning rate $10^{-4}$ (decayed at epoch 80 and 150), batch size 128, 200 epochs, and $\lambda=10^{-2}$.

For a more accurate evaluation, we fine-tune on the training set of ImageNet-100 \cite{cao2017hashnet} an Inception-V3 network that was pre-trained on ImageNet \cite{imagenet_cvpr09}. We fine-tune the Inception-V3 with the SGD optimizer, 50 epochs, batch size 128, and initial learning rate $10^{-4}$. The fine-tuned Inception-V3 is then used to compute Intra-FID, FID, and IS.

\subsection{More details of the efficiency analysis}

\begin{figure}[h]
	\centering
	\subfloat[][Storage Usage (GB) for DRE-F-SP+RS]{\includegraphics[width=0.45\textwidth]{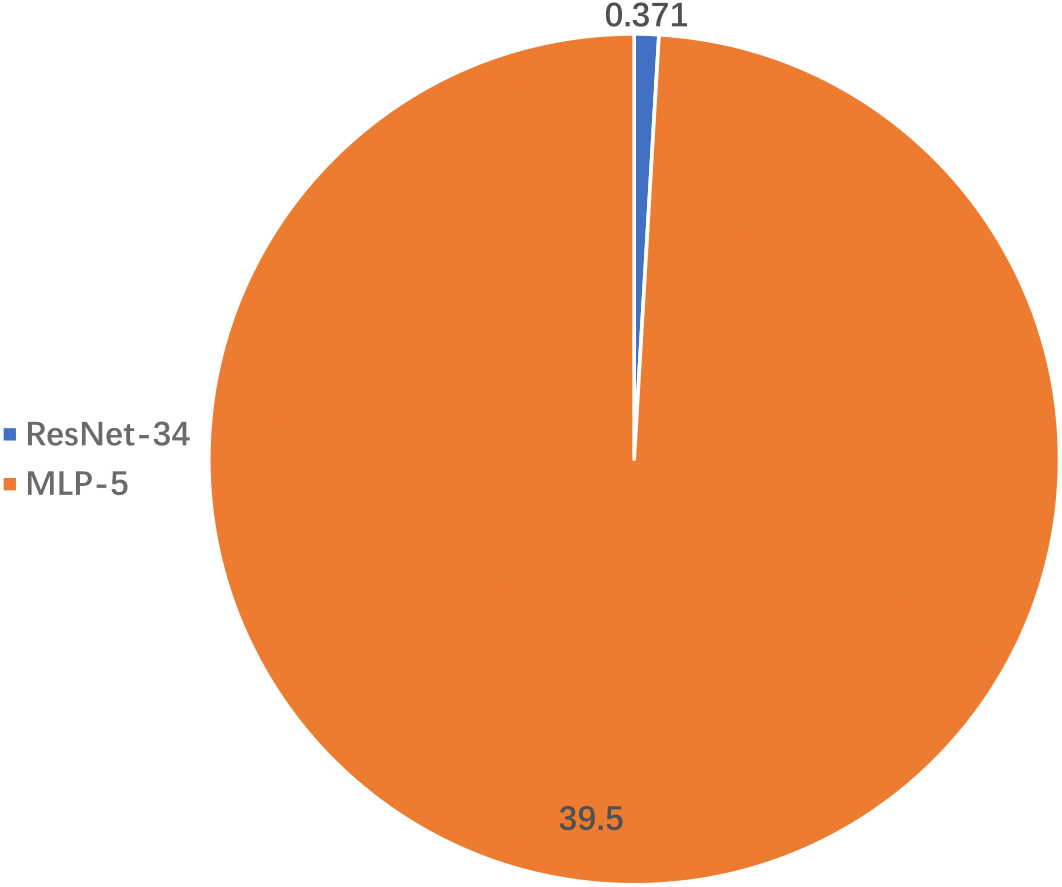}\label{fig:ImageNet-100_storage_DRE}}\quad
	\subfloat[][Storage Usage (GB) for cDR-RS]{\includegraphics[width=0.45\textwidth]{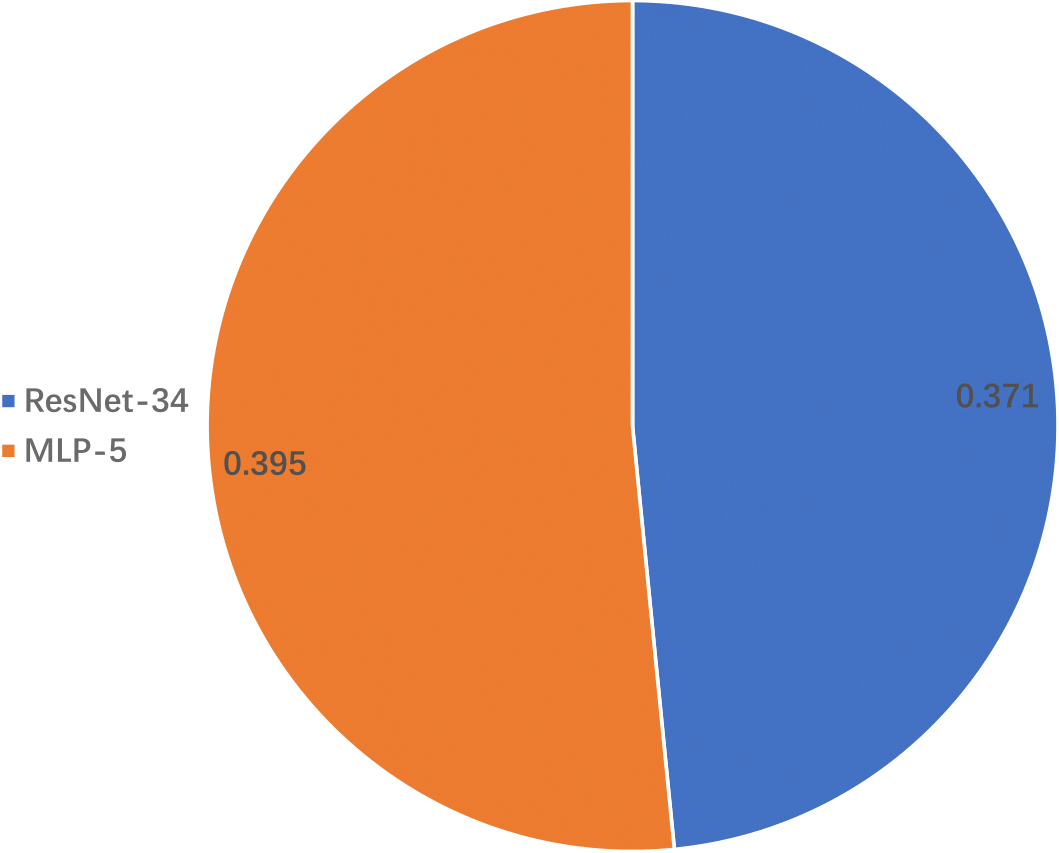}\label{fig:ImageNet-100_storage_cDRE}}
	\\
	\subfloat[][Training Time (hours) for DRE-F-SP+RS]{\includegraphics[width=0.45\textwidth]{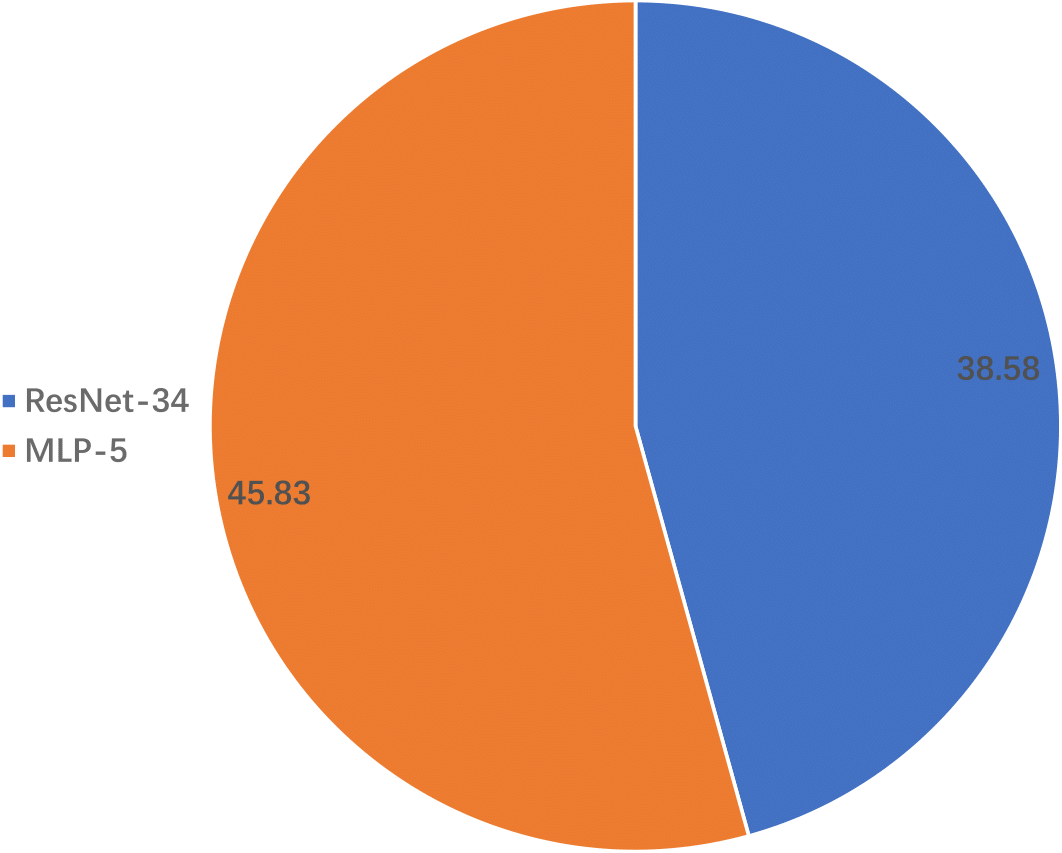}\label{fig:ImageNet-100_time_DRE}}\quad
	\subfloat[][Training Time (hours) for cDR-RS]{\includegraphics[width=0.45\textwidth]{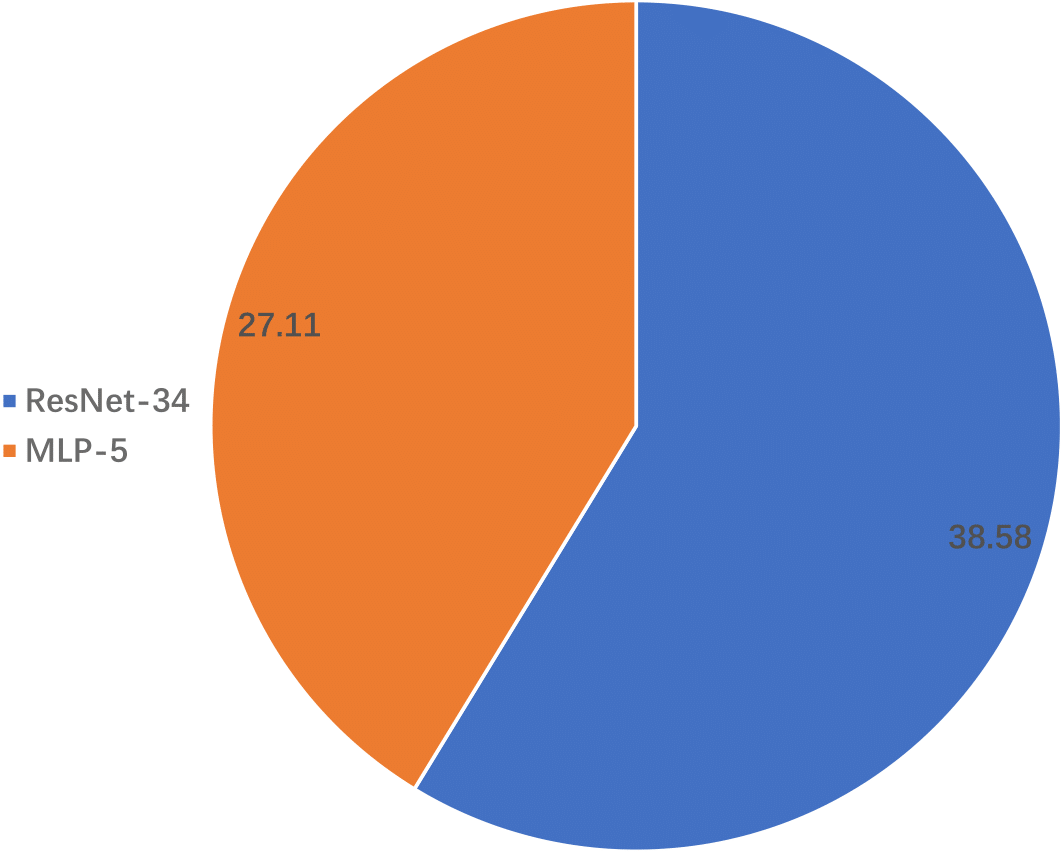}\label{fig:ImageNet-100_time_cDRE}}
	\\
	\subfloat[][Total Implementation Time (hours) for DRE-F-SP+RS]{\includegraphics[width=0.45\textwidth]{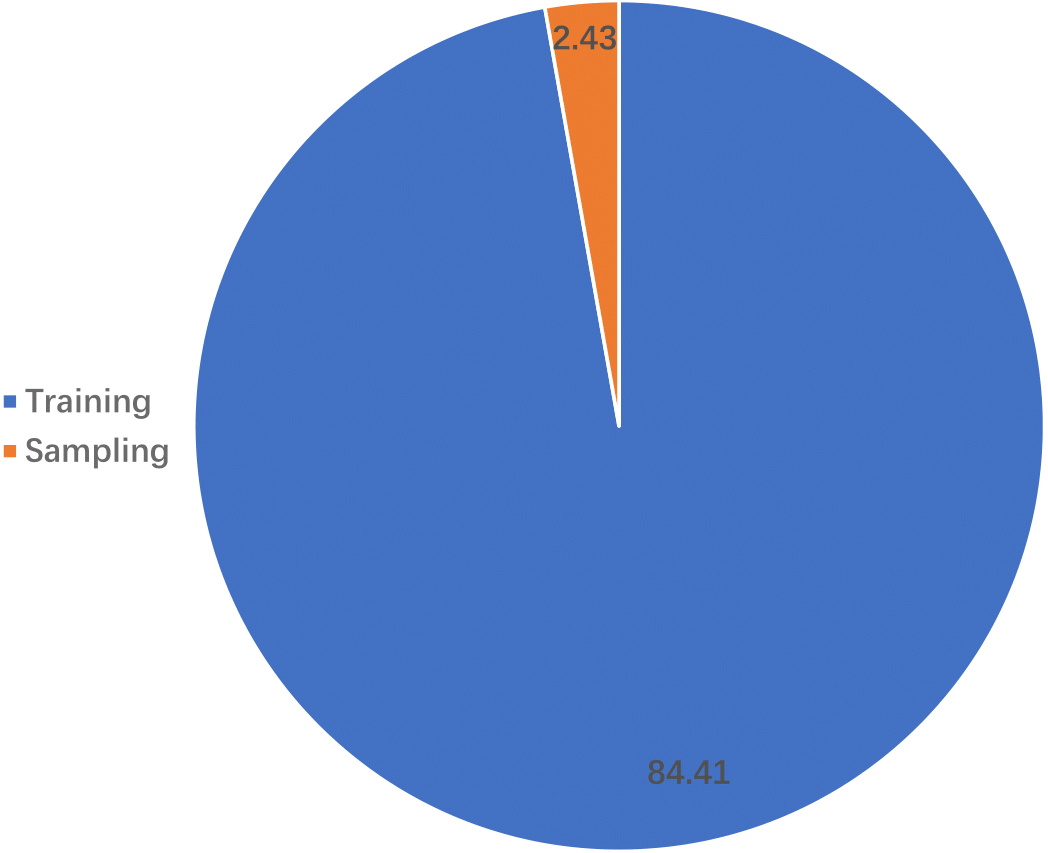}\label{fig:ImageNet-100_total_time_DRE}}\quad
	\subfloat[][Total Implementation Time (hours) for cDR-RS]{\includegraphics[width=0.45\textwidth]{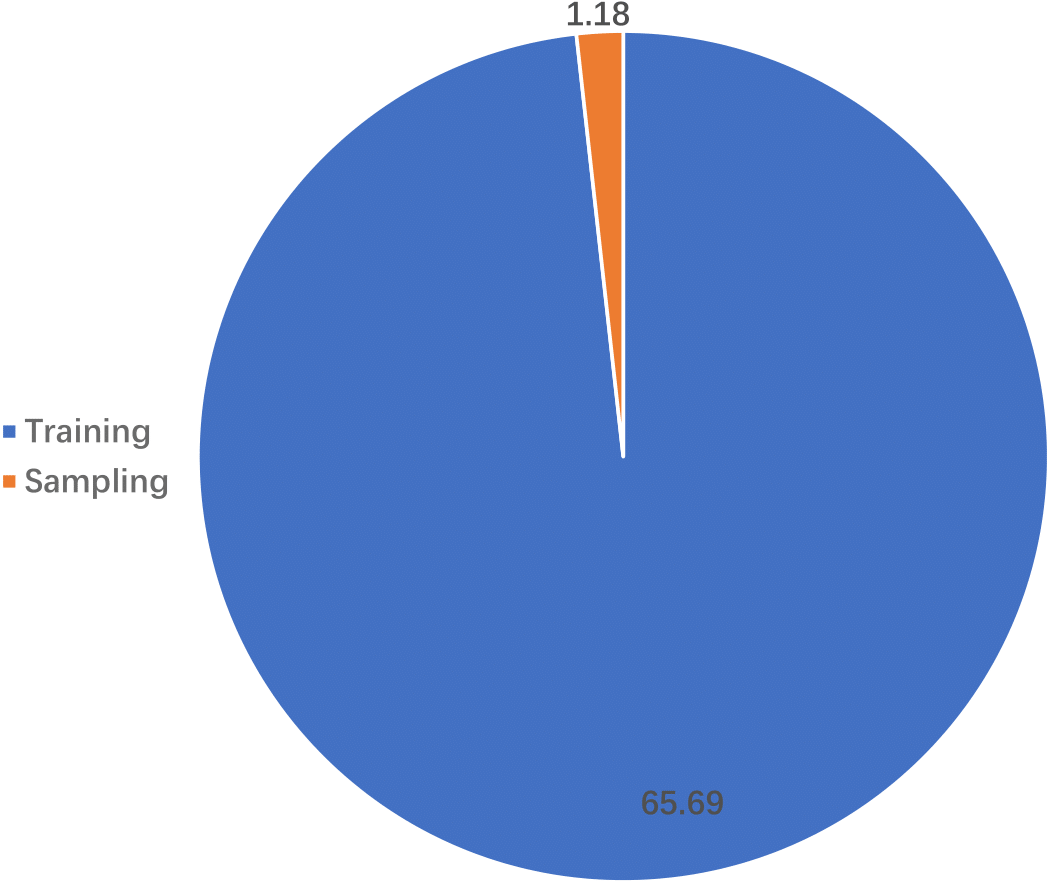}\label{fig:ImageNet-100_total_time_cDRE}}
	\caption{Pie charts for the efficiency analysis of DRE-F-SP+RS and cDR-RS on ImageNet-100.}
	\label{fig:ImageNet-100_efficiency_analysis_pie_charts}
\end{figure}

In order to compare the efficiency of candidate sampling methods, we summarize their storage usage and training and sampling time in \Cref{tab:ImageNet-100_efficiency_analysis_supp}, based on which we plot \cref{fig:cover_fig}. Some pie charts are also plotted to show more detailed storage usage and training time for DRE-F-SP+RS and cDR-RS in \cref{fig:ImageNet-100_efficiency_analysis_pie_charts}. \textbf{Please note that, the storage usage here, is the overall storage usage consumed by all models (e.g., the discriminator network in DRS and density ratio models in cDR-RS) in each sampling method except the generator network of cGANs.} For DRE-F-SP+RS, the 100 MLP-5 models take a lot of disk space (almost 40 GB), making it less efficient in subsampling class-conditional GANs with many classes. We use two NVIDIA V100 GPUs (32GB) for this analysis. 

\begin{table}[htbp]
	\centering
	\caption{Efficiency analysis of different sampling methods on ImageNet-100 based on Two NVIDA V100. For DRE-F-SP+RS and cDR-RS, the training time includes the time spent on the ResNet-34 training and the MLP-5 network training.}
	\begin{adjustbox}{width=0.7\textwidth}
		\begin{tabular}{lcccc}
			\toprule
			\textbf{Methods} & \textbf{ \begin{tabular}[c]{@{}c@{}} Total storage \\ usage (GB) \end{tabular} } & \textbf{ \begin{tabular}[c]{@{}c@{}} Total training \\ time (hours) \end{tabular} } & \textbf{ \begin{tabular}[c]{@{}c@{}} Total sampling \\ time (hours) \end{tabular} } & \textbf{ \begin{tabular}[c]{@{}c@{}} Total implementation \\ time (hours) \end{tabular} } \\
			\midrule
			Collab & 0.13  & 5.58  & 5.23  & 10.81 \\
			DRS   & 0.13  & 1.47  & 0.75  & 2.22 \\
			DDLS  & 0.13     & 0     & 218.05 & 218.05 \\
			DRE-F-SP+RS & 38.94 & 84.41 & 2.43  & 86.84 \\
			cRS-RS & 0.75  & 65.69  & 1.18  & 66.87 \\
			\bottomrule
		\end{tabular}%
	\end{adjustbox}
	\label{tab:ImageNet-100_efficiency_analysis_supp}%
\end{table}%

\section{More Details of Experiments on UTKFace}\label{supp:details_of_utkface}

\subsection{Setups of training, sampling, and evaluation}

We follow \cite{ding2021ccgan, ding2020continuous} to implement CcGANs (SVDL+ILI) with the SNGAN architecture. The detailed setups can be found in \cite{ding2021ccgan, ding2020continuous} or our codes. Please note that, due to randomness in training, our reported evaluation results of CcGANs, such as Intra-FID, are slightly different from those in \cite{ding2021ccgan, ding2020continuous}. 

To implement Collab, we conduct discriminator shaping for 3000 iterations for CcGAN. We do the refinement 16 times in a middle layer of the generator network of CcGAN. The step size of refinement is set as 0.5. 

To implement DRS, we fine-tune the discriminator of CcGAN for 2000 iterations with batch size 128.  

To implement DDLS, we run the Langevin dynamics procedure for CcGAN for each age with step size $10^{-4}$ up to 400 iterations. More iterations will make the sampling process too time-consuming.

To implement DRE-F-SP+RS, we first train the specially designed SAE on the training set for 200 epochs with the SGD optimizer, initial learning rate 0.01 (decayed every 50 epochs with factor 0.1), weight decay $10^{-4}$, batch size 256, and sparsity parameter $\lambda=10^{-3}$. The network architecture of this SAE is shown in \Cref{tab:SparseAE_encoder_64x64}, \Cref{tab:SparseAE_decoder_64x64}, and \Cref{tab:SparseAE_labelPred_64x64}. Sixty MLP-5 models for modeling the density ratio function at each age are trained on the training set with the Adam optimizer \cite{kingma2014adam}, initial learning rate $10^{-4}$ (decayed at epoch 100 and 250), batch size 256, 400 epochs, and $\lambda=10^{-2}$. The network architecture of MLP-5 is similar to \Cref{tab:cifar10_MLP5}. Similar to above experiments, when implementing DRE-F-SP+RS, we use more epochs than cDR-RS does to let all density ratio models converge. Nevertheless, in this experiment, there are still many density ratio models do not converge, which may be one reason of the failure of DRE-F-SP+RS in subsampling CcGANs.

To implement cDR-RS, we use the specially designed SAE in the implementation of DRE-F-SP+RS to extract features from images. The MLP-5 to model the conditional density ratio function is similar to \Cref{tab:cifar10_cMLP5}. It is trained with the Adam optimizer \cite{kingma2014adam}, initial learning rate $10^{-4}$ (decayed at epoch 80 and 150), batch size 256, 200 epochs, and $\lambda=10^{-2}$. The $\zeta$ in the filtering scheme of cDR-RS is set to be $0.1$. 

The models for the computation of Intra-FID, NIQE, Diversity, and Label Score are consistent with those used by \cite{ding2021ccgan, ding2020continuous}. We quote and rephrase the definitions of these metrics in \cite{ding2021ccgan, ding2020continuous} as follows.
\begin{itemize}
	\item \textbf{Intra-FID} \cite{miyato2018cgans}: ``We take Intra-FID as the overall score to evaluate the quality of fake images and we prefer the small Intra-FID score. At each evaluation angle, we compute the FID \cite{heusel2017gans} between real images and 1000 fake images in terms of the bottleneck feature of the pre-trained AE."
	
	\item \textbf{NIQE} \cite{mittal2012making}: ``NIQE is used to evaluate the visual quality of fake images with the real images as the reference and we prefer the small NIQE score."
	
	\item \textbf{Diversity}: ``Diversity is used to evaluate the intra-label diversity and the larger the better." In UTKFace, there are 6 races. At each age, we ask a pre-trained classification-oriented ResNet-34 to predict the races of the 1000 fake images and an entropy is computed based on these predicted races. The Diversity score is the average of the entropies computed on all ages. 
	
	\item \textbf{Label Score}: ``Label Score is used to evaluate the label consistency and the smaller the better." We ask the pre-trained regression-oriented ResNet-34 to predict the ages of all fake images and the predicted ages are then compared with the conditioning ages. The Label Score is defined as the average absolute distance between the predicted ages and conditioning ages over all fake images, which is equivalent to the Mean Absolute Error (MAE). 
\end{itemize}
Please refer to the official implementation of CcGANs at \url{https://github.com/UBCDingXin/improved_CcGAN} for more details. 
\begin{table}[h]
	\centering
	\caption{The architecture of the encoder in the sparse autoencoder for extracting features from $64\times 64$ RGB images. In convolutional (Conv) operations, $ch$ denotes the number of channels, $k/s/p$ denote kernel size, stride and number of padding, respectively. }
		\begin{tabular}{c}
			\toprule
			Input: an RGB image $\bm{x}\in \mathbbm{R}^{64\times64\times3}$. \\ 
			\hline
			Conv ($ch\rightarrow64$, $k4/s2/p1$), BN, ReLU \\\hline
			Conv ($ch\rightarrow64$, $k3/s1/p1$), BN, ReLU \\\hline
			Conv ($ch\rightarrow64$, $k4/s2/p1$), BN, ReLU \\\hline
			Conv ($ch\rightarrow128$, $k3/s1/p1$), BN, ReLU \\\hline
			Conv ($ch\rightarrow128$, $k4/s2/p1$), BN, ReLU \\\hline
			Conv ($ch\rightarrow256$, $k3/s1/p1$), BN, ReLU \\\hline
			Conv ($ch\rightarrow256$, $k4/s2/p1$), BN, ReLU \\\hline
			Conv ($ch\rightarrow256$, $k3/s1/p1$), BN, ReLU \\ \hline
			Flatten: $64\times 4 \times 4\times 4 \rightarrow 4096$ \\ \hline
			fc$\rightarrow 12288$, ReLU \\\hline
			Output: extracted sparse features $\bm{h}\in \mathbbm{R}^{12288}$
			\\ \bottomrule
		\end{tabular}%
	\label{tab:SparseAE_encoder_64x64}%
\end{table}%

\begin{table}[h]
	\centering
	\caption{The architecture of the decoder in the sparse autoencoder for reconstructing $64\times64$ input images from extracted features. In transposed-convolutional (ConvT) operations, $ch$ denotes the number of channels, $k/s/p$ denote kernel size, stride and number of padding, respectively. }
		\begin{tabular}{c}
			\toprule
			Input: extracted sparse features $\bm{h}\in \mathbbm{R}^{12288}$ 
			\\ from Table~\ref{tab:SparseAE_encoder_64x64}. \\
			\hline
			fc$\rightarrow 4096$, BN, ReLU \\\hline
			Reshape: $4096\rightarrow 64\times 4\times 4 \times 4$\\ \hline
			ConvT ($ch\rightarrow256$, $k4/s2/p1$), BN, ReLU \\\hline
			ConvT ($ch\rightarrow128$, $k3/s1/p1$), BN, ReLU \\\hline
			ConvT ($ch\rightarrow128$, $k4/s2/p1$), BN, ReLU \\\hline
			ConvT ($ch\rightarrow64$, $k3/s1/p1$), BN, ReLU \\\hline
			ConvT ($ch\rightarrow64$, $k4/s2/p1$), BN, ReLU \\\hline
			ConvT ($ch\rightarrow64$, $k3/s1/p1$), BN, ReLU \\\hline
			ConvT ($ch\rightarrow64$, $k4/s2/p1$), BN, ReLU \\\hline
			ConvT ($ch\rightarrow64$, $k3/s1/p1$), BN, ReLU \\\hline
			ConvT ($ch\rightarrow3$, $k1/s1/p0$), Tanh
			\\ \hline
			Output: a reconstructed image $\bm{x}\in \mathbbm{R}^{64\times 64\times 3}$
			\\ \bottomrule
		\end{tabular}%
	\label{tab:SparseAE_decoder_64x64}%
\end{table}%

\begin{table}[h]
	\centering
	\caption{The architecture of the label prediction branch in the sparse autoencoder for $64\times64$ images.}
		\begin{tabular}{c}
			\toprule
			Input: extracted sparse features $\bm{h}\in \mathbbm{R}^{12288}$ \\
			from Table~\ref{tab:SparseAE_encoder_64x64}. \\
			\hline
			fc$\rightarrow 1024$, BN, ReLU \\\hline
			fc$\rightarrow 512$, BN, ReLU \\\hline
			fc$\rightarrow 256$, BN, ReLU \\\hline
			fc$\rightarrow 1$, ReLU \\
			\hline
			Output: the predicted label $\hat{y}$
			\\ \bottomrule
		\end{tabular}%
	\label{tab:SparseAE_labelPred_64x64}%
\end{table}%

\subsection{More details of the efficiency analysis}
\begin{figure}[h]
	\centering
	\subfloat[][Storage Usage (MB) for DRE-F-SP+RS]{\includegraphics[width=0.45\textwidth]{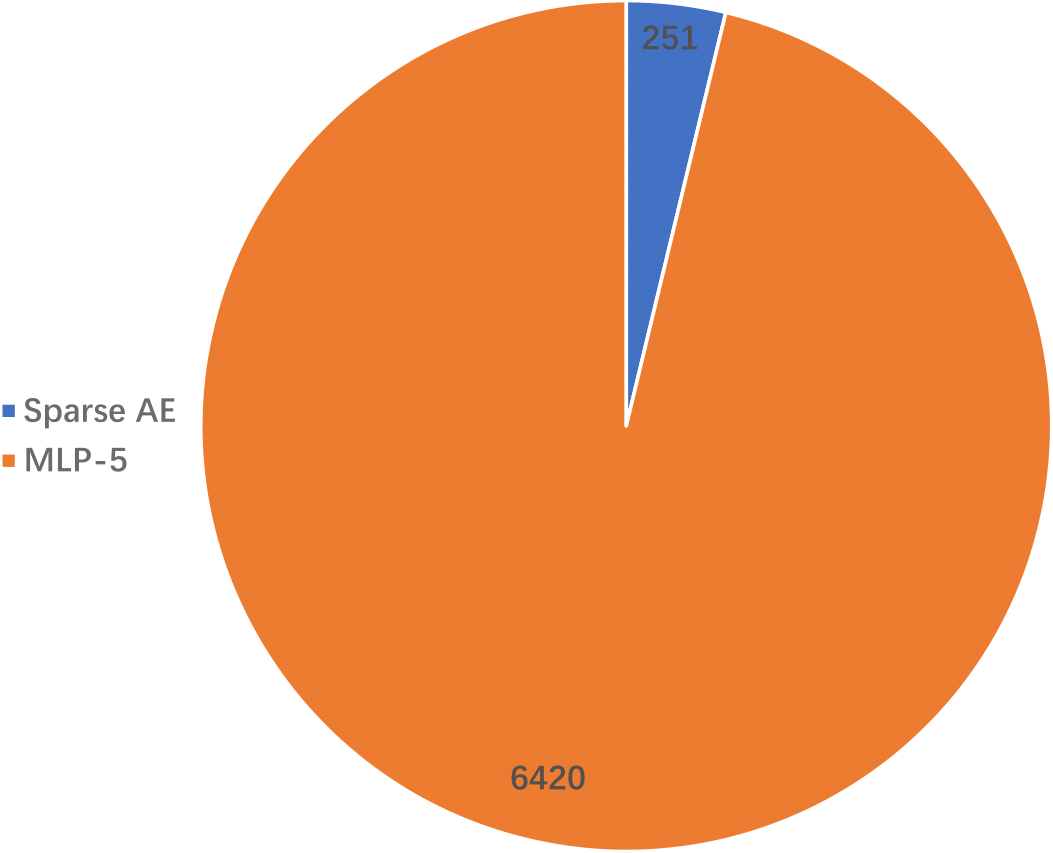}\label{fig:UTKFace_storage_DRE}}\quad
	\subfloat[][Storage Usage (MB) for cDR-RS]{\includegraphics[width=0.45\textwidth]{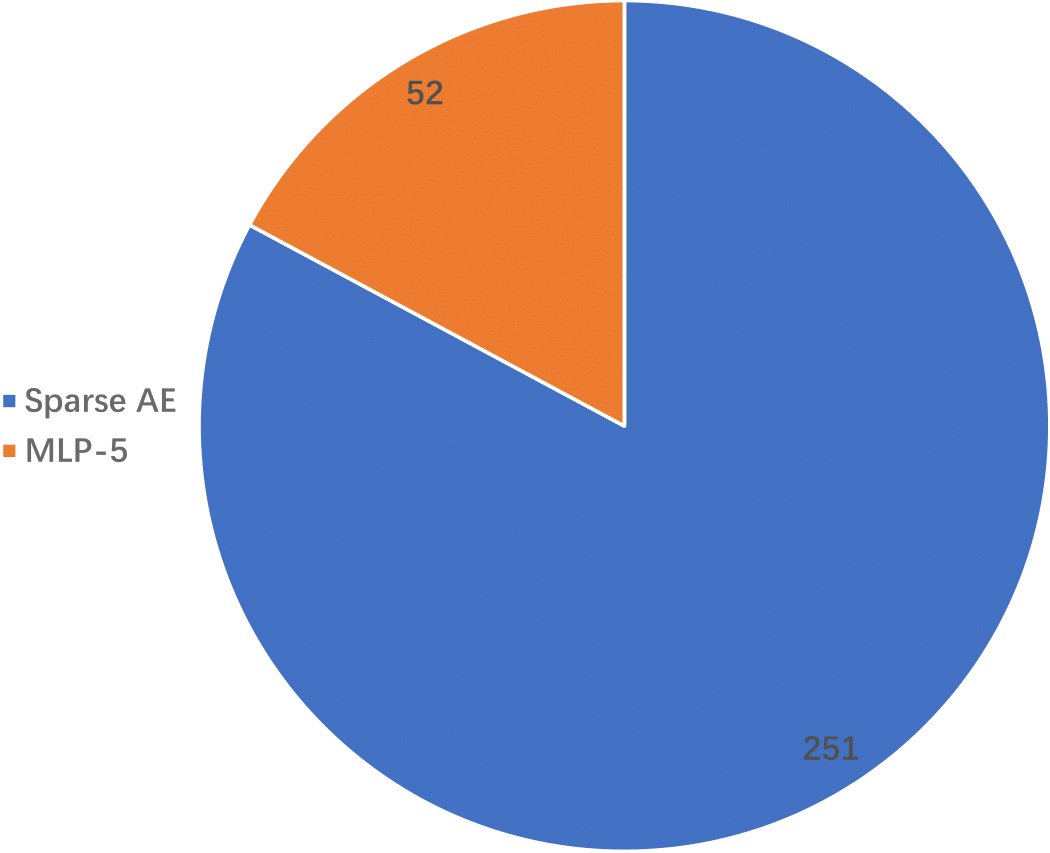}\label{fig:UTKFace_storage_cDRE}}
	\\
	\subfloat[][Training Time (hours) for DRE-F-SP+RS]{\includegraphics[width=0.45\textwidth]{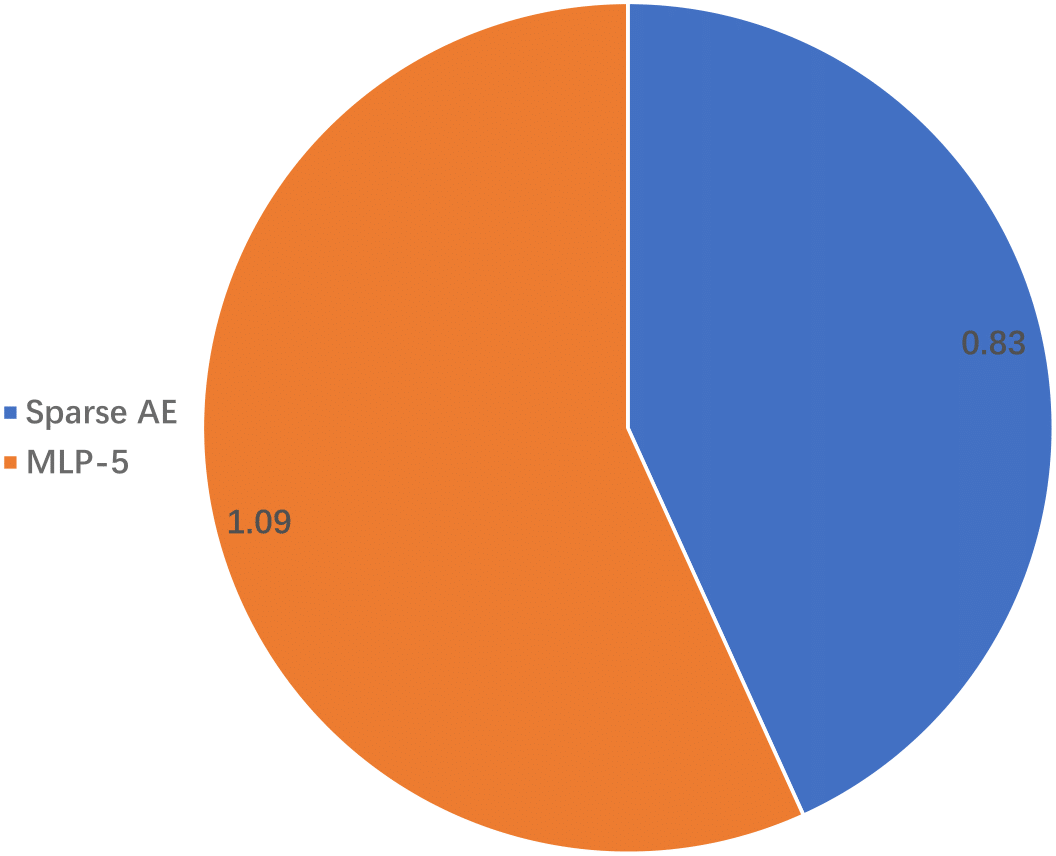}\label{fig:UTKFace_time_DRE}}\quad
	\subfloat[][Training Time (hours) for cDR-RS]{\includegraphics[width=0.45\textwidth]{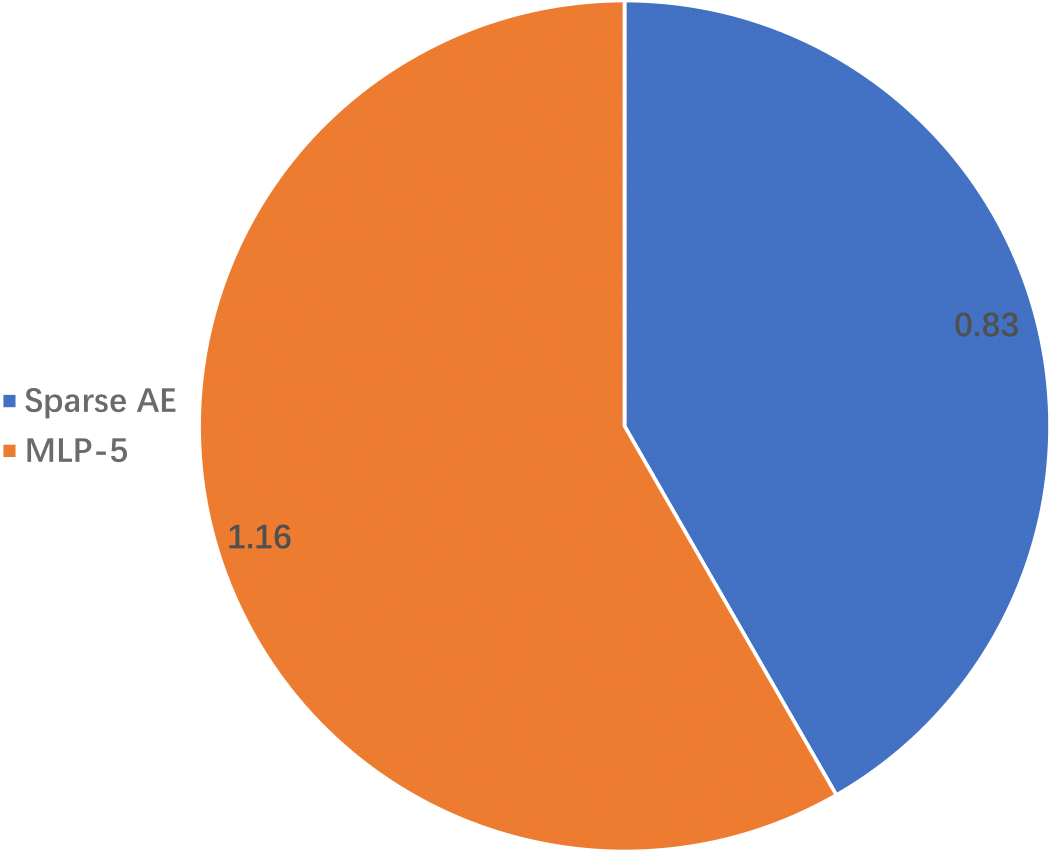}\label{fig:UTKFace_time_cDRE}}
	\\
	\subfloat[][Total Implementation Time (hours) for DRE-F-SP+RS]{\includegraphics[width=0.45\textwidth]{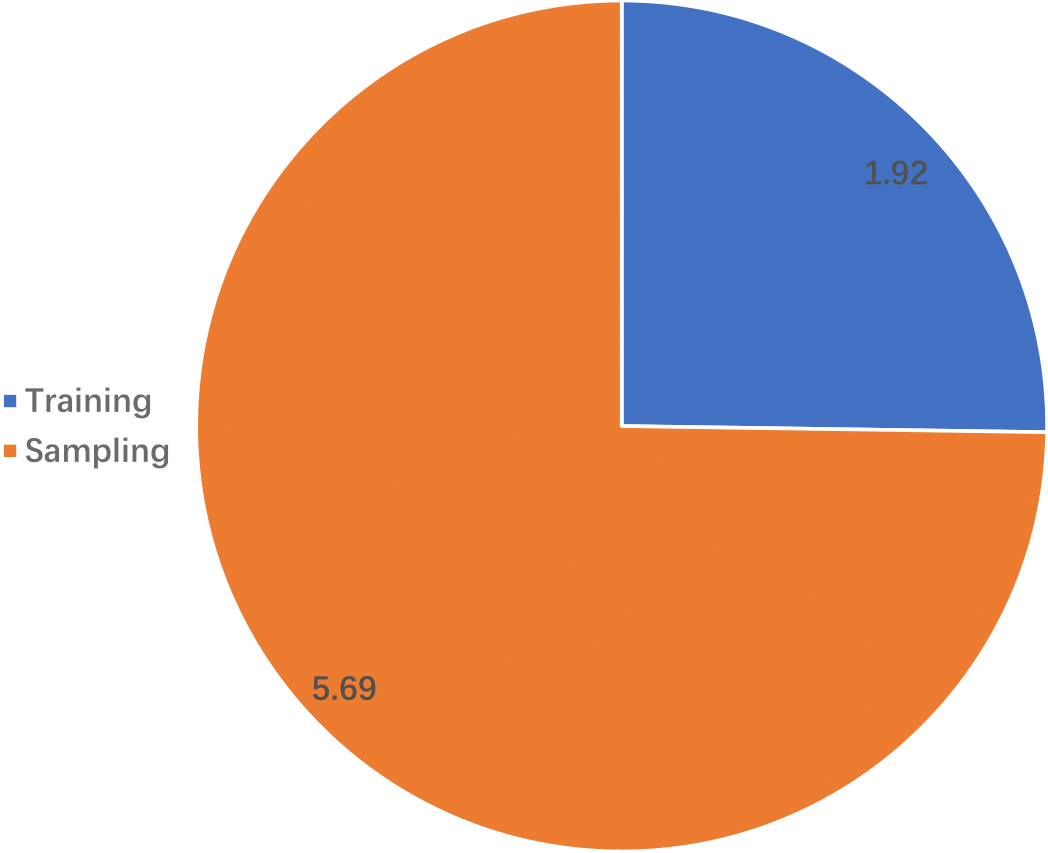}\label{fig:UTKFace_total_time_DRE}}\quad
	\subfloat[][Total Implementation Time (hours) for cDR-RS]{\includegraphics[width=0.45\textwidth]{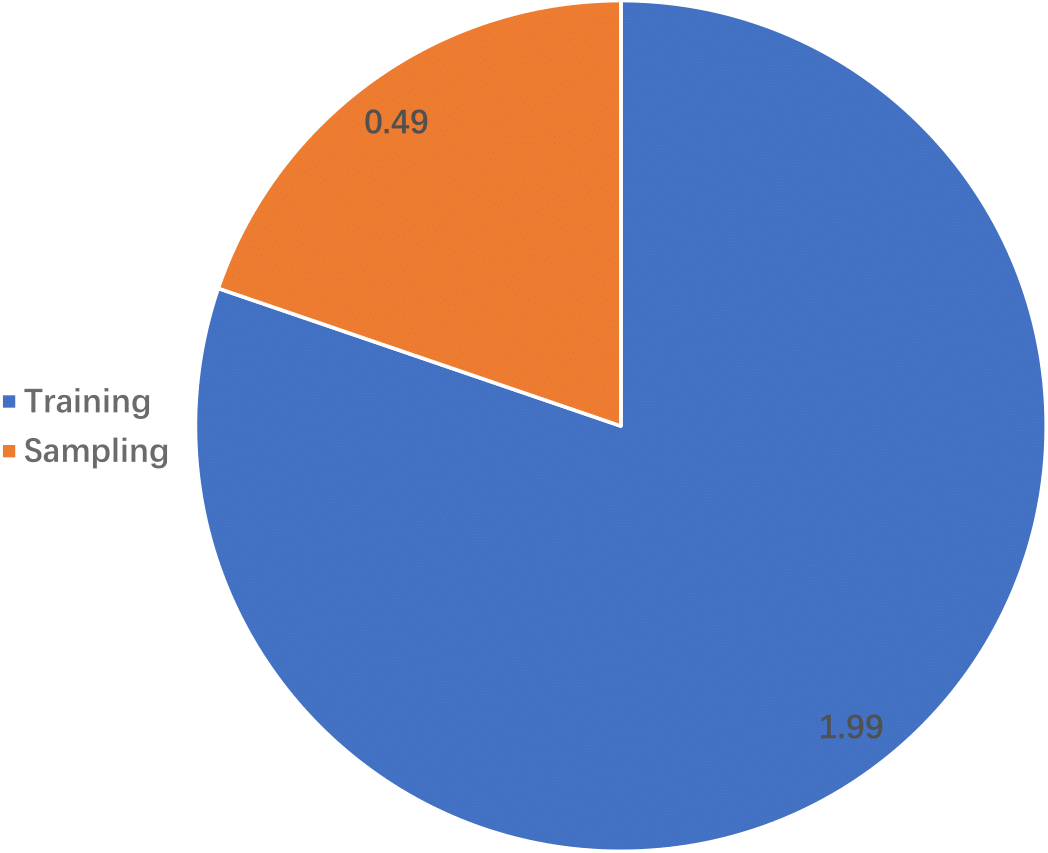}\label{fig:UTKFace_total_time_cDRE}}
	\caption{Pie charts for the efficiency analysis of DRE-F-SP+RS and cDR-RS on UTKFace.}
	\label{fig:UTKFace_efficiency_analysis_pie_charts}
\end{figure}

Similar to the ImageNet-100 experiment, we analyze the efficiency of all candidate methods, and summarize the results in \Cref{tab:UTKFace_efficiency_analysis_supp} and \cref{fig:UTKFace_efficiency_analysis_pie_charts}. \textbf{Please note that, the storage usage here, is the overall storage usage consumed by all models (e.g., the discriminator network in DRS and density ratio models in cDR-RS) in each sampling method except the generator network of cGANs.} The 60 MLP-5 models take a lot of disk space for DRE-F-SP+RS. We use one NVIDIA V100 GPUs (16GB) for this analysis.

\begin{table}[htbp]
	\centering
	\caption{Efficiency analysis of different sampling methods on UTKFace based on One NVIDA V100. For DRE-F-SP+RS and cDR-RS, the training time includes the time spent on the SAE training and the MLP-5 network training.}
	\begin{adjustbox}{width=0.7\textwidth}
		\begin{tabular}{lccccc}
			\toprule
			\textbf{Methods} & \textbf{ \begin{tabular}[c]{@{}c@{}} Total storage \\ usage (MB) \end{tabular} } & \textbf{ \begin{tabular}[c]{@{}c@{}} Total training \\ time (hours) \end{tabular} } & \textbf{ \begin{tabular}[c]{@{}c@{}} Total sampling \\ time (hours) \end{tabular} } & \textbf{ \begin{tabular}[c]{@{}c@{}} Total implementation \\ time (hours) \end{tabular} } \\
			\midrule
			Collab & 82.8  & 1.05  & 0.27  & 1.32 \\
			DRS   & 82.8  & 0.16  & 0.13  & 0.29 \\
			DDLS  & 82.8  & 0  & 11.24 & 11.24 \\
			DRE-F-SP+RS & 6,671 & 1.92 & 5.69  & 7.61 \\
			cRS-RS & 303  & 1.99  & 0.49  & 2.48 \\
			\bottomrule
		\end{tabular}%
	\end{adjustbox}
	\label{tab:UTKFace_efficiency_analysis_supp}%
\end{table}%

\section{More Details of Experiments on RC-49}\label{supp:details_of_rc49}

Similar to the UTKFace experiment, we follow \cite{ding2021ccgan, ding2020continuous} to implement CcGANs (SVDL+ILI) with the SNGAN architecture. The detailed setups can be found in \cite{ding2021ccgan, ding2020continuous} or our codes. Please note that, due to randomness in training, our reported evaluation results of CcGAN (SVDL+ILI), such as Intra-FID, are slightly different from those in \cite{ding2021ccgan, ding2020continuous}. 

To implement Collab, we conduct discriminator shaping for 3000 iterations for CcGAN. We do the refinement 16 times in a middle layer of the generator network of CcGAN. The step size of refinement is set as 0.5. 

To implement DRS, we fine-tune the discriminator of CcGAN for 1000 iterations with batch size 128.  

DDLS is not implemented on RC-49 due to a too long sampling time.

DRE-F-SP+RS is not applicable to this scenario, where we need to generate images conditional on labels that are unseen in the training phase.

To implement cDR-RS, first train the specially designed SAE on the training set for 200 epochs with the SGD optimizer, initial learning rate 0.01 (decayed every 50 epochs with factor 0.1), weight decay $10^{-4}$, batch size 256, and sparsity parameter $\lambda=10^{-3}$. The MLP-5 to model the conditional density ratio function is similar to \Cref{tab:cifar10_cMLP5}. It is trained with the Adam optimizer \cite{kingma2014adam}, initial learning rate $10^{-4}$ (decayed at epoch 80 and 150), batch size 256, 200 epochs, and $\lambda=10^{-3}$. The $\zeta$ in the filtering scheme of cDR-RS is set to be $0.13$. 

In the testing phase, we generate 179,800 fake images (200 per angle) from each candidate method over 899 distinct labels. This experiment adopts four evaluation metrics---(i) Intra-FID \cite{miyato2018cgans} is an overall image quality metric; (ii) \textit{Naturalness Image Quality Evaluator} (NIQE) \cite{mittal2012making} evaluates the visual quality of fake images. Please note again that the visual quality is only one aspect of image quality; (iii) \textit{Diversity} measures the diversity of fake images; and (iv) \textit{Label Score} (LS) evaluates label consistency. 

Specifically, the four metrics are computed as follows. (i) For the Intra-FID index, at each of the 899 angles ($0.1^{\circ}-89.9^{\circ}$), we compute the FID \cite{heusel2017gans} value between 49 real images and 200 fake images in terms of the bottleneck feature of the pre-trained autoencoder. The Intra-FID score is the average FID over all 899 evaluation angles. (ii) For the NIQE index, firstly we fit an NIQE model with the 49 real rendered chair images at each of the 899 angles which gives 899 NIQE models. We then compute an average NIQE score for each evaluated angle using the NIQE model at that angle. Finally, we report the average of the 899 average NIQE scores over the 899 yaw angles. The block size and the sharpness threshold are set to 8 and 0.1 respectively in this experiments. We employ the built-in NIQE library in \texttt{MATLAB}. (iii) For the Diversity index, at each evaluation angle, firstly we use a pretrained classification-oriented ResNet-34 to predict the chair types (49 types in total) of these 200 fake images. Then, an entropy value can be computed based on the chair type predictions at this angle. Finally, the Diversity index is defined as the average of the entropies at all 899 angles. (iv) For the Label Score index, at each evaluation angle, firstly we ask a pretrained regression-oriented ResNet-34 to predict the yaw angles of all fake image samples and the predicted angles are then compared with the assigned angles. The Label Score value is defined as the average absolute distance between the predicted angles and assigned angles over all fake images.

\end{document}